\documentclass[final]{cvpr}

\usepackage{times}
\usepackage{epsfig}
\usepackage[utf8]{inputenc} 
\usepackage[T1]{fontenc}    
\usepackage{url}            
\usepackage{booktabs}       
\usepackage{amsfonts}       
\usepackage{nicefrac}       
\usepackage{microtype}      
\usepackage{graphicx}
\usepackage{subfigure}

\usepackage{times}
\usepackage{epsfig}
\usepackage{amsmath}
\usepackage{amssymb}
\usepackage{textgreek}
\usepackage{tabularx}
\usepackage{multirow}
\usepackage{adjustbox}
\usepackage{xspace}
\usepackage{xcolor}

\graphicspath{{figures/}}

\usepackage[pagebackref=true,breaklinks=true,colorlinks,bookmarks=false]{hyperref}

\newcommand{\R}{\mathbb{R}}
\newcommand{\SO}{\ensuremath{\mathrm{SO}}}

\newcommand{\InteriorNetA}{InteriorNet-A\xspace}
\newcommand{\InteriorNetB}{InteriorNet-B\xspace}
\newcommand{\MatterportA}{Matterport-A\xspace}
\newcommand{\MatterportB}{Matterport-B\xspace}

\newenvironment{packed_enum}{
\begin{enumerate}
  \setlength{\itemsep}{1pt}
  \setlength{\parskip}{2pt}
  \setlength{\parsep}{0pt}
}{\end{enumerate}}

\def\cvprPaperID{3629} 
\def\confYear{CVPR 2021}

\begin{document}


\title{Wide-Baseline Relative Camera Pose Estimation with Directional Learning}
\author{%
  Kefan Chen\thanks{Work done while Kefan was a member of the Google AI Residency program (g.co/airesidency).} \\
  Google Research\\
  {\tt\small chenkefan950518@gmail.com} \\
   \and
   Noah Snavely \\
  Google Research\\
   {\tt\small snavely@google.com} \\
   \and
   Ameesh Makadia \\
  Google Research\\
   {\tt\small makadia@google.com} 
}

\pagenumbering{gobble}
\maketitle

\begin{abstract}
Modern deep learning techniques that regress the relative camera pose between two images have difficulty dealing with challenging scenarios, such as large camera motions resulting in occlusions and significant changes in perspective that leave little overlap between images.  These models continue to struggle even with the benefit of large supervised training datasets.  To address the limitations of these models, we take inspiration from techniques that show regressing keypoint locations in 2D and 3D can be improved by estimating a discrete distribution over keypoint locations. Analogously, in this paper we explore improving camera pose regression by instead predicting a discrete distribution over camera poses. To realize this idea, we introduce DirectionNet, which estimates discrete distributions over the 5D relative pose space using a novel parameterization to make the estimation problem tractable. Specifically, DirectionNet factorizes relative camera pose, specified by a 3D rotation and a translation direction, into a set of 3D direction vectors. Since 3D directions can be identified with points on the sphere, DirectionNet estimates discrete distributions on the sphere as its output. We evaluate our model on challenging synthetic and real pose estimation datasets constructed from Matterport3D and InteriorNet. Promising results show a near 50\% reduction in error over direct regression methods. 
Code will be available at {\footnotesize\url{https://arthurchen0518.github.io/DirectionNet}}.

\end{abstract}

\section{Introduction}
Estimating the relative pose between two images is fundamental to many applications in computer vision such as 3D reconstruction, stereo rectification, and camera localization~\cite{mvg03book}.
For calibrated cameras, relative pose is synonymous with the essential matrix, which encapsulates the projective geometry relating two views.
Prevailing approaches recover the global model from corresponding points~\cite{LonguetHiggins81,hartley8pt,nister03cvpr} within an iterative robust model fitting process~\cite{ransac,lmeds}.  Recent progress has introduced deep-learned modules that can replace components of this classic pipeline~\cite{yi18cvpr,ranftl2018eccv,Dusmanu2019CVPR,DeTone_2018_CVPR_Workshops,sarlin20superglue,lfnet,probst2019cvpr,brachmann17cvpr}. While this class of techniques has been extensively analyzed~\cite{ransacsurvey}, well-known failure cases include where feature detection or matching is difficult, such as low image overlap, large changes in scale or perspective, or scenes with insufficient or repeated textures.

In these cases, it is natural to consider if supervised deep learning can address this basic task of essential matrix estimation, given its success in a variety of challenging computer vision problems. Specifically, can we train a deep neural network to represent the complex function that directly maps image pairs to their relative camera pose? Such a model would provide an appealing alternative to formulations that are sensitive to correspondence estimation performance.

Unfortunately, evidence suggests designing regression models for pose estimation is challenging, and in fact finding a parameterization of the motion groups effective in deep learning models is still an active research topic~\cite{mahendran17cvprw,zhou19cvpr,peretroukhin20rss}. Not surprisingly, the initial works exploring relative pose regression (e.g.~\cite{melekhov17acivs,poursaeed2018eccv}) are not conclusively successful in the difficult scenarios described above.

In this work we introduce a novel deep learning model for relative pose estimation, focused on the challenging wide-baseline case. Conceptually, our model generates a discrete probability distribution over relative poses, and the final pose estimate is taken as the expectation of this distribution. Our method is inspired by works that show estimating a discrete distribution, or a heatmap over a quantized output space, consistently outperforms direct regression to the continuous output space. This observation has arisen in various applications, including estimating 2D and 3D keypoint locations~\cite{sun2017integral,luvizon19humanpose,suwajanakorn2018discovery} and estimating periodic angles~\cite{rcnn3d}.

However, it is currently unclear if this idea translates directly to complex higher-dimensional output spaces. Relative camera pose lives in a five dimensional space, so predicting a discrete distribution would require $O(N^5)$ storage in the output space alone~\cite{makadia07ijcv}! Given that representing such large spaces is currently intractable for neural networks at any reasonable resolution, how can we effectively apply this concept to the relative pose problem?

To this end, we introduce a novel formulation that builds upon the idea of estimating discrete probability distributions on the 5D relative pose space. We propose two key components to execute this idea effectively:
\begin{packed_enum}
\item A parameterization of the motion space that factorizes poses as a set of 3D direction vectors. A non-parametric differentiable projection step can map these directions to their closest pose.
\item \textit{DirectionNet}, a convolutional encoder-decoder model for predicting sets of 3D direction vectors. The network outputs discrete distributions on the sphere $S^2$, the expected values of which produce direction vectors.
\end{packed_enum}
\textit{Our core contributions are in recognizing that incorporating a dense structured output in the form of a discrete probability distribution can improve wide-baseline relative pose estimation, and in introducing a technique to execute this idea efficiently}. The attributes of this approach that help make it effective include (1) DirectionNet is fully-convolutional and as such does not utilize any fully-connected regression layers, and (2) it allows for additional supervision as both the dense distribution and final estimated pose can be supervised.

DirectionNet is deployed in two stages, wherein the relative rotation is estimated first, followed by the relative translation direction. This allows us to de-rotate the input images after the first stage, which reduces the complexity of the translation estimation task.

DirectionNet is evaluated on two difficult wide-baseline pose estimation datasets created from the synthetic images in InteriorNet \cite{InteriorNet18}, and the real images in Matterport3D \cite{Matterport3D}.
DirectionNet consistently outperforms direct regression approaches (for the same pose representation as well as numerous alternatives), as well as classic feature-based approaches. This illustrates the effectiveness of estimating discrete probability distributions as an alternative to direct regression, even for a complex problem like relative pose estimation.  Furthermore, these results validate that a supervised data-driven approach for wide-baseline pose estimation can succeed in cases that are extreme for traditional methods.

\section{Related Work}
Feature matching--based methods are still prevalent for the relative pose problem, yet suffer under large motions that yield unreliable correspondence (see~\cite{ransacsurvey} for a survey). Recent works deploy deep learning in subproblems such as feature detection~\cite{sarlin20superglue,Dusmanu2019CVPR}, filtering or reweighting outliers~\cite{yi18cvpr,ranftl2018eccv}. Differentiable versions of consensus methods like RANSAC have also been proposed~\cite{probst2019cvpr,brachmann17cvpr}. These still rely on sufficiently accurate matches, an uncertain prospect in wide-baseline settings.

Many deep regression methods have addressed 3D object pose recovery from a single image~\cite{MousavianCVPR17,su15iccv,vpsKpsTulsianiM15,li18eccv,sundermeyer2018eccv}. For our task of relative pose estimation, \cite{melekhov17acivs} adopts a Siamese convolutional regression model to directly estimate relative camera pose from two images. \cite{en18eccvw} proposes a relative pose layer atop Siamese camera localization towers. \cite{poursaeed2018eccv} introduces a model for uncalibrated cameras that regresses to a fundamental matrix via an intermediate representation of camera intrinsics, rotation, and translation. Related to these efforts, \cite{detone16rssw} and \cite{nguyen18iral} proposed deep convolutional networks for homography estimation. Despite targeting various tasks, most of the above methods share a common architecture design---a convolutional network culminating with fully connected layers for regression. A related problem to ours is camera re-localization which estimates pose from a single image in a \emph{known} scene~\cite{kendall15iccv,Fragoso2017PatchCF}. Our setting is quite different as we try to recover the \textit{relative} camera pose from two images in a previously unvisited scene.

Deep learning for ego-motion estimation or visual odometry is an active area which includes supervised methods such as \cite{ummenhoferZUMID16} (depth supervision), and many self-supervised methods (see~\cite{chen2020survey} for a survey). In general, these systems make design choices specific for small-baselines and video sequences, such as using frame-to-frame image reconstruction losses~\cite{zhou2017unsupervised,geonet,d3vo}, or training with more than two frames~\cite{deepvo,dso,zhou2017unsupervised}. In contrast, our   focus is on learning to estimate relative pose for wide-baseline image pairs.

Probabilistic deep models have been used to capture uncertainty in pose predictions. In~\cite{poserbpf, manhardt2018} they estimate the distribution of 6D object poses to tackle shape symmetries and ambiguities, while~\cite{deepdirectstat2018, Gilitschenski2020Deep} uses directional statistics to model object rotations and regress the parameters of the probability distributions. \cite{Campbell_2019_CVPR} adopts mixtures of von Mises-Fisher and quasi-Projected Normal distributions to represent point sets. The multi-headed approach in~\cite{hydranet} combines multiple predictions into a mean pose and associated covariance matrix. In contrast to most of these techniques, ours is a discrete representation and is not tied to any choice of parametric probability distribution model.
\begin{figure*}[t]
  \begin{center}
    \includegraphics[width=0.99\textwidth]{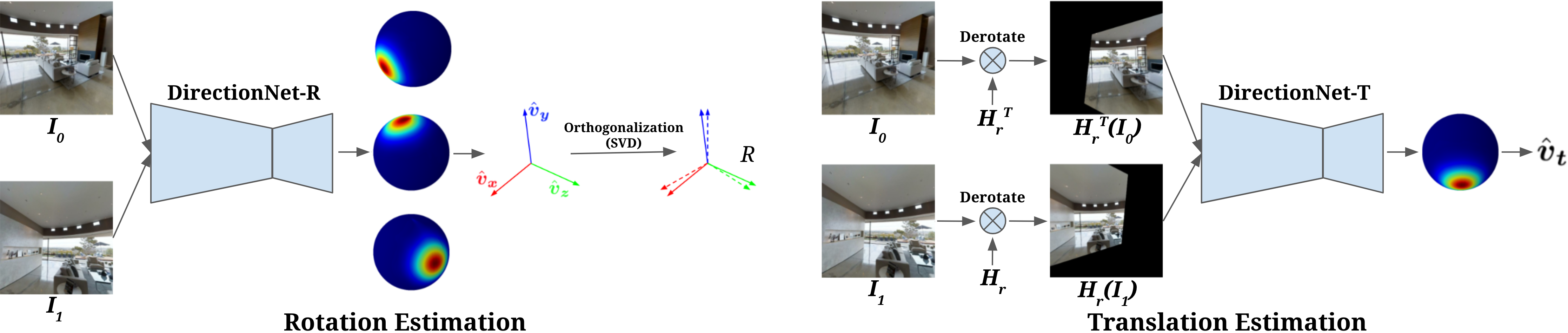}
  \end{center}
  \caption{\textbf{Relative pose estimation.} At the core of our method is the DirectionNet, which maps a source image $I_0$ and a target image $I_1$ to a number of directional probability distributions over the 2-sphere, shown here as color-coded spheres. We convert the distributions to vectors by finding their expected values. The rotation matrix $R$ is approximated by orthogonal Procrustes from three estimated unit vectors $(\hat{v}_{x}, \hat{v}_{y}, \hat{v}_{z})$. As an alternative, DirectionNet-R could generate two directional vectors and $R$ could be determined by  Gram-Schmidt orthogonalization. To facilitate estimating the translation $\hat{v}_{t}$, we derotate the input images by applying the homography introduced in Sec~\ref{eq:Homography}, yielding the transformed input images $H_{r}^{T}(I_0)$ and $H_r(I_1)$ where $r$ is half-rotation of the estimated camera rotation $R$.} 
  \label{fig:Model}
\end{figure*}

\section{Method}
We now outline our method for estimating relative pose from image pairs. The relative pose between two views is specified by a 3D rotation $R\in \SO(3)$ ($R \in \mathbb{R}^{3\times3}$, $R^\top R=I$, $\mathrm{det}(R)=1$), and a translation $t \in \R^3$. Without additional assumptions,
the relative translation can only be recovered up to a scale factor, so we adopt the normalization $\left\Vert t \right\Vert = 1$; equivalently, $t$ is restricted to the 2-sphere, $t \in S^2$. Our task is to estimate the relative pose in $\SO(3)\times S^2$.

In principle, we desire a model that estimates a discrete distribution over the pose space $\SO(3)\times S^2$. This requires discretizing this five-dimensional space at a reasonable resolution, which is computationally infeasible in deep networks. Instead, our approach presents a novel parameterization for relative poses along with simplifying assumptions to make the task tractable.

\subsection{Directional parameterization of relative pose}
Parameterizing $\SO(3)\times S^2$ requires a choice for $\SO(3)$.
We choose a simple over-parameterization of $\SO(3)$, splitting the rotation matrix $R \in \SO(3)$ into its component vectors $R=\left[x\;y\;z\right]$, where each component is itself a unit vector in $\mathbb{R}^3$. The relative pose between images, $(R,t) \in \SO(3)\times S^2$, can then be specified by the four direction vectors $x, y, z, t \in S^2$.
See Section~\ref{sec:analysis} for a discussion and justification of our parameterization.
\subsection{Estimating 3D directions}
The relative pose task is now one of estimating a set of direction vectors ${x, y, z, t \in S^2}$. Making the simplifying assumption of independence of these vectors, our approach is to (1) predict a probability distribution over the space of possible directions ($S^2$) for each vector, and (2) extract the direction prediction from each distribution.
It is important to note that representing functions on the sphere and integrating them requires careful consideration in the context of neural networks where the data representation is restricted to regular grids. The details of our approach are below.

\paragraph{Spherical distributions.}
We represent functions on $S^2$ with a 2D equirectangular projection indexed by spherical coordinates $(\theta_i, \phi_j)$. Our discretization follows~\cite{kostelec2008}: $\theta_i$ is the angle of colatitude ($0 \leq i < H$, $\theta_i=\frac{(2i+1)\pi}{2H}$), $\phi_j$ is the azimuth ($0 \leq j < W$, $\phi_j=\frac{2\pi j}{W}$), and $H\times W$ is the grid resolution.
Let $u(\theta_i, \phi_j)$ denote the unnormalized output of a network, and $f(x) = \ln(1 + e^x)$ be the softplus function. We can map $u$ to a probability distribution with a spherical normalization:
\begin{equation}
    P(\theta_i, \phi_j) = \frac{f(u(\theta_i, \phi_j))}{\sum_{i=0}^{H-1}\sum_{j=0}^{W-1}f(u(\theta_i, \phi_j))\sin(\theta_i)}\label{eq:distribution},
\end{equation}
where the $\mathrm{sin}(\theta_i)$ in the normalizing term comes from the area element on $S^2$.

\paragraph{Spherical expectation.}
The $\operatorname{argmax}$ operator identifies the most probable direction from a distribution, but is not differentiable and its precision is limited by the grid resolution. An alternative is the Fr\'{e}chet mean~\cite{afsari2011}, which is appealing since it defines a spherical centroid using the natural metric (geodesics). However, it requires a non-convex optimization and the solution is not necessarily unique. Instead, we choose the alternative of taking the expected value of the distribution. 
In the continuous case, we define the expected value of a random variable $X$ on $S^2$ with PDF $p_{X}$ as $\mathrm{E}[X] = \int_{\rho\in S^2} \rho\cdot p_{X}(\rho) d\rho$. 
In the discretization of the sphere introduced above, this becomes
\begin{equation}
    \mathrm{E}[X] = \sum_{i=0}^{H-1}\sum_{j=0}^{W-1}\rho(\theta_i, \phi_j)P(\theta_i, \phi_j)\sin(\theta_i),\label{eq:expectation}
\end{equation} 
where $\rho(\theta_i, \phi_j)$ is the 3D unit vector corresponding to spherical point $(\theta_i, \phi_j)$. The expected value $v=\mathrm{E}[X], v \in \mathbb{R}^3$ can be projected to the sphere in a straightforward manner: $\hat{v} = \frac{v}{\lVert v \rVert}$.\footnote{An alternative formulation to Equations~\ref{eq:distribution} and \ref{eq:expectation} would include the traditional $\operatorname{soft-argmax}$ operator adapted to the sphere. This would reinterpret the network output $u$ as log-probabilities and require $f(x)=e^x$ in Eq.~\ref{eq:distribution}.}

\begin{figure*}[tp]
\subfigure[DirectionNet architecture]{\label{fig:NetworkArchitecture}\includegraphics[width=0.78\textwidth]{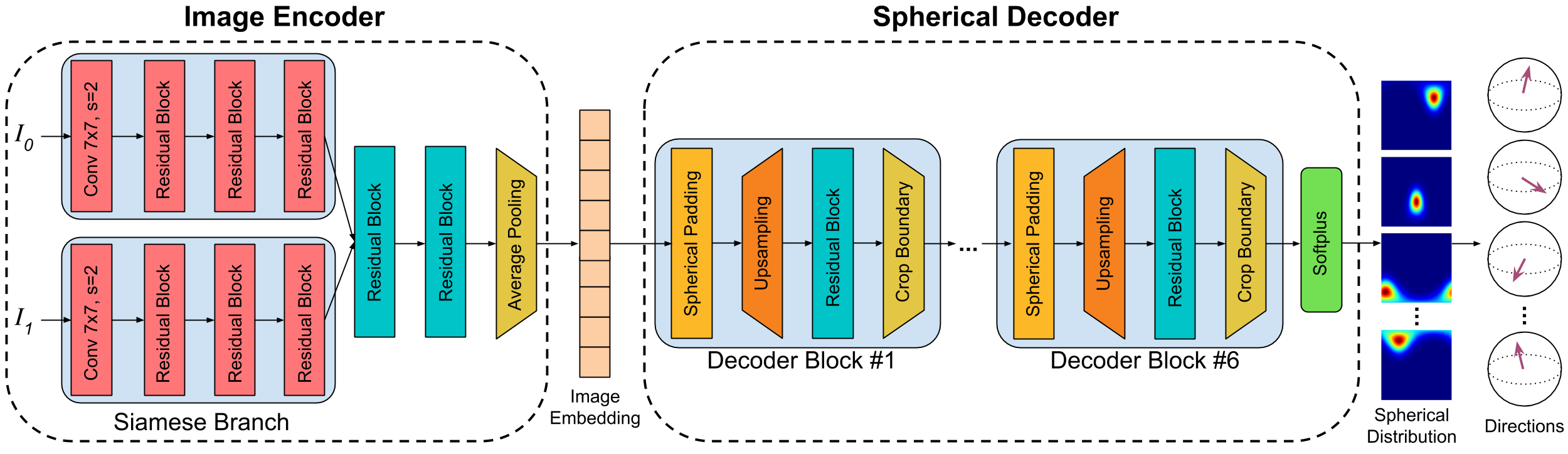}}
\subfigure[Spherical padding]{\label{fig:SphericalPadding}\includegraphics[width=0.2\textwidth]{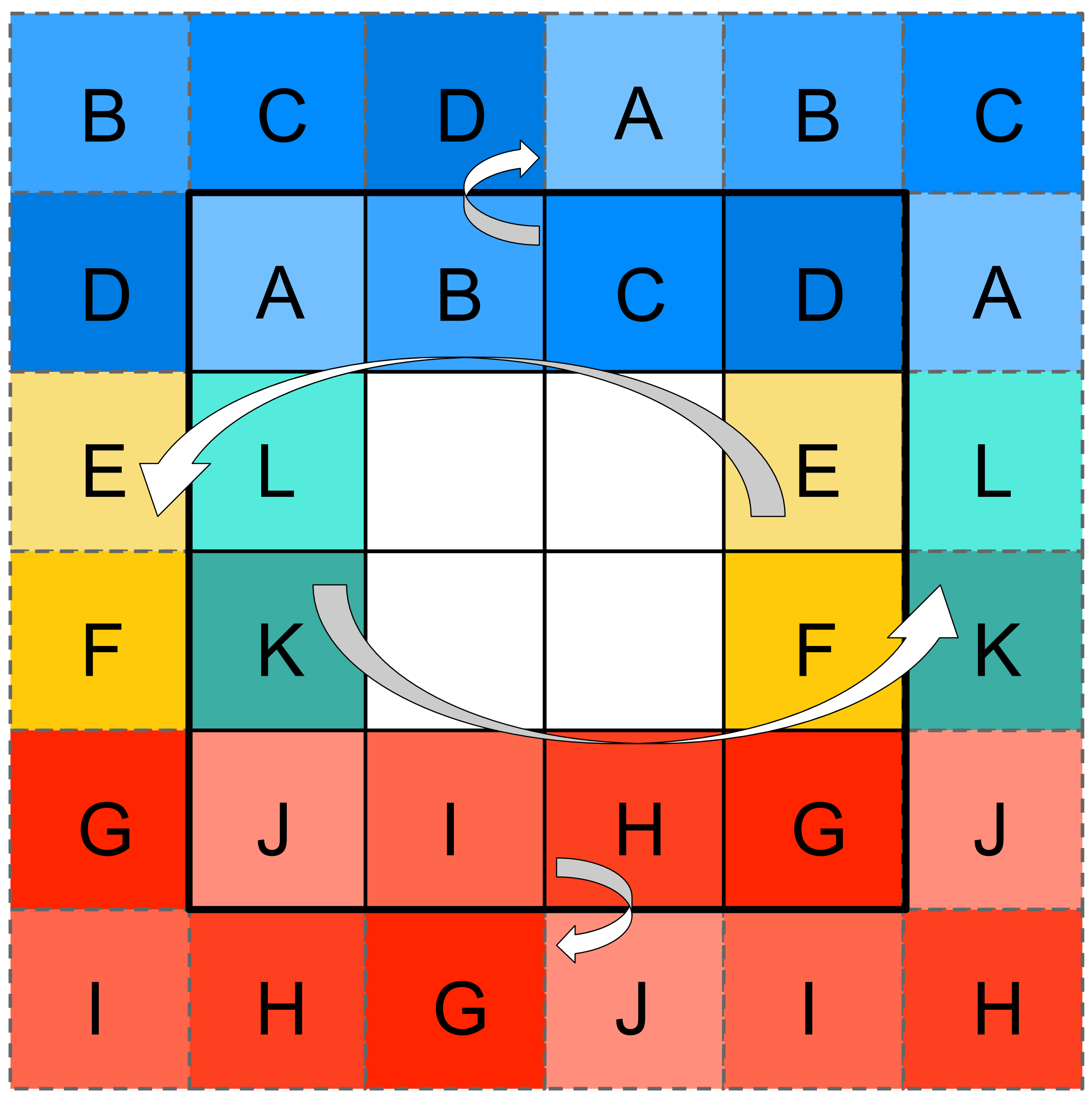}}
\caption{(a) The image encoder generates embeddings from a pair of input images, and the spherical decoder transforms and upsamples these embeddings to produce probability distributions over $S^2$, which are represented with equirectangular maps.
(b) Spherical padding ensures the boundary pixels reflect the correct neighbors on the sphere. In this example, a 4$\times$4 grid is padded to size 6$\times$6. The corresponding labeled squares illustrate the padding process. See appendix \ref{apppadding} for further clarification.
}
\end{figure*}

\subsection{DirectionNet for relative pose estimation}\label{secprocrustes}
DirectionNet maps image pairs to sets of unit direction vectors following the steps outlined above. We adopt an encoder-decoder style architecture that learns a cross-domain mapping from two images to a spherical representation (see Figure~\ref{fig:NetworkArchitecture}). We describe two ways that DirectionNet can be instantiated for relative pose estimation.

\paragraph{The SVD variation.} Here DirectionNet will produce four vectors $\{\hat{v}_{x}, \hat{v}_{y}, \hat{v}_{z}, \hat{v}_{t}\}$ which are the predictions of the directional pose components $\{x, y, z, t\}$.  Due to the simplifying assumptions earlier, in general $\left[\hat{v}_{x},\;\hat{v}_{y},\;\hat{v}_{z}\right] \notin \SO(3)$ as orthogonality is not guaranteed. However, we can project $M=\left[\hat{v}_{x},\;\hat{v}_{y},\;\hat{v}_{z}\right]$ to $\SO(3)$ with orthogonal Procrustes~\cite{procrustes}, which is a differentiable procedure~\cite{suwajanakorn2018discovery} and provides the optimal projection to $\SO(3)$ by Frobenius norm:
\begin{equation}
    R = U\mathrm{diag}(1,1,\det(UV^T))V^T,\label{eq:orthogonalization}
\end{equation}
where $U\Sigma V^T = M$ is the SVD of M~\cite{procrustes}.

\paragraph{The Gram-Schmidt variation.} For $\left[\hat{v}_{x},\;\hat{v}_{y},\;\hat{v}_{z}\right] \in \SO(3)$, the last column $\hat{v}_{z}$ is fully constrained: $\hat{v}_{z} = \hat{v}_{x} \times \hat{v}_{y}$. Thus, we only need DirectionNet to produce three vectors $\{\hat{v}_{x}, \hat{v}_{y}, \hat{v}_{t}\}$ and obtain a rotation by a partial Gram-Schmidt projection of $\hat{v}_{x}$ and $\hat{v}_{y}$ as in~\cite{zhou19cvpr}.

Both the SVD and Gram-Schmidt projections have been shown recently to reach state-of-the-art performance for different 3D rotation estimation tasks, especially when predicting arbitrary (large) rotations. See \cite{levinson20neurips} and \cite{zhou19cvpr} for the analysis. Although overall performance between the two is similar, SVD is shown to be slightly more effective, with a possible explanation being that SVD finds a least-squares projection while Gram-Schmidt is greedy.  Our experimental findings confirm the analysis.

\subsection{Two-stage model with derotation}
Intuitively, we expect that learning camera translation should be easiest when the data never exhibit rotational motion. Hence, we propose estimating camera pose sequentially: (1) A DirectionNet (denoted DirectionNet-R) estimates the relative rotation between input images, (2) this rotation is used to \emph{derotate} the input images, and (3) a second DirectionNet (denoted DirectionNet-T) predicts translations from the rotation-free image pair. Figure~\ref{fig:Model} illustrates the stages of this process.

However, when the relative rotation between the cameras is large, derotating one image relative to the other could result in projecting most of the scene outside the camera's field of view. 
To limit this effect we find that it is helpful to project both input images to an intermediate frame using half of the estimated rotation with a larger FoV and proportionally increased resolution. The derotation implementation involves a homography transformation
$H_r=K^\prime r^\top K^{-1} \label{eq:Homography}$,
where $K$ is the matrix of camera intrinsics for the input image, $K^\prime$ is the intrinsics for the desired derotated image, and $r$ is the half-rotation of the estimated rotation $R$. Setting $K^\prime = K$ maintains the field of view (FoV) and resolution of the input image after derotation. We use ${H_r}^T$ and $H_r$ to project input images $I_0$ and $I_1$, respectively, onto the ``middle'' frame. The output FoV and resolution are controllable parameters of the model. To mitigate the effect of rotation errors on translation prediction, we developed a rotation augmentation scheme with random perturbations (see appendix \ref{apptrainingdetails}).

\subsection{Network architecture}
The DirectionNet architecture is illustrated in Figure~\ref{fig:NetworkArchitecture}. Notably, we eschew the skip connections common to similar convolutional architectures (\cite{newell2016stacked,RonnebergerFB15}) since they would not reflect the correct spatial associations between the planar and spherical topologies (observed in \cite{esteves19icml}).

The \textbf{image encoder} embeds image pairs into $\mathbb{R}^{512}$ using a Siamese architecture. A Siamese branch consists of a $7{\times}7$, stride-2 convolution followed by a series of residual blocks\footnote{All residual blocks consist of two bottleneck blocks~\cite{resnet} with $3{\times}3$ convolutions, batch normalization, and leaky ReLU pre-activations.}, each of which downsamples by 2. The outputs are concatenated in the channel dimension, and after two more residual blocks, a global average pooling produces the embedding.

The \textbf{spherical decoder} maps embeddings to spherical distributions by repeatedly applying bilinear upsampling and residual blocks. We use spherical padding before upsampling to ensure adjacent pixels at the boundaries reflect the correct neighbors on the sphere (see Figure~\ref{fig:SphericalPadding}). The final outputs, of size $64{\times}64{\times}k$, are interpreted as $k$ spherical distributions, and are subsequently mapped to $k$ direction vectors following equations~\ref{eq:distribution} and~\ref{eq:expectation}. In total, a single DirectionNet contains ${\sim}9$M parameters. Note that we could also consider a Spherical CNN decoder~\cite{s.2018spherical,esteves18eccv}. While such models are equivariant to 3D rotations, the benefit in our setting would be limited since the image encoder is not rotation equivariant. Hence, we opt for the computational efficiency of 2D convolutions.

\subsection{Loss terms and model training}
In this section, we describe the loss terms and training strategy for our model. We let $P(\theta_i, \phi_j)$ denote a single distribution generated by DirectionNet, and let its corresponding ground truth distribution be $P^*(\theta_i, \phi_j)$. The ground truth distributions are constructed using the von Mises-Fisher distribution as described in Sec.~\ref{ImplementationDetails}. With a slight abuse of notation we will denote with $E[P]$ the expected value of a distribution and is computed according to equation~\ref{eq:expectation}. One appealing property of DirectionNet is that both the output direction vectors as well as their corresponding dense probability distributions can be supervised.

The \textbf{direction loss} is the negative cosine similarity between two 3D vectors:
\begin{equation}\small
    L_{\nearrow}(p_1, p_2) = - \frac{p_1^T p_2}{\lVert p_1\rVert \lVert p_2\rVert}.
\end{equation}
We introduce two loss terms to supervise distributions. First, the \textbf{distribution loss} provides dense supervision on the equirectangular distribution grid:
\begin{equation}  
\small
L_{D}(P_1, P_2) = \frac{1}{HW}\sum_{i=0}^{H-1}\sum_{j=0}^{W-1}(P_1(\theta_i, \phi_j)-P_2(\theta_i, \phi_j))^2\sin(\theta_i),
\end{equation}
Second, the \textbf{spread loss} penalizes the spherical ``variance''~\cite{mardia1975} to encourage unimodal and concentrated distributions:
 \begin{equation}\small
     L_{\sigma}(p) = 1-\lVert p \rVert.
\end{equation}
The full loss for a single predicted direction vector combines the three individual losses:
\begin{eqnarray}\small
    L(P, P^*) & = & L_{\nearrow}(E[P], E[P^*]) + \lambda_DL_{D}(P, P^*) + \nonumber\\
    & &  \lambda_{\sigma}L_{\sigma}(E[P])). \label{eq:loss}
\end{eqnarray} 

See appendix \ref{appresultsdiscuss} for an analysis of the individual loss terms. By far the most impactful loss on performance is the distribution loss $L_{D}$.

We train our two-stage pipeline sequentially. We first train DirectionNet-R, which takes the source and target image pair ($I_0$, $I_1$) as input and produces three direction vectors $\left[\hat{v}_{x}\;\hat{v}_{y}\;\hat{v}_{z}\right]$, which are mapped onto $\SO(3)$. The complete training loss for DirectionNet-R is the sum of individual losses for the three estimated directions $L_R = L(P_x, P_x^*) + L(P_y, P^*_y) + L(P_z, P^*_z)$. After training, DirectionNet-R is frozen and its predictions are used for derotating the inputs of DirectionNet-T. DirectionNet-T takes the input pair ($H_{r}^T(I_0)$, $H_{r}(I_1)$) and outputs a translation direction $\hat{v}_{t}$. Since the translation is represented with a single unit vector, DirectionNet-T is trained with the loss $L_T = L(P_t, P^*_t)$.

\begin{table*}[t!]
  \centering
  \newcommand{\B}[1]{\textbf{#1}}
  \begin{adjustbox}{width=1.\linewidth}
  \begin{tabular}{llcccccccccccccccc}
    \toprule
    \multicolumn{2}{l}{} & \multicolumn{7}{c}{\MatterportA} & \phantom{a} & \phantom{a}& \multicolumn{7}{c}{\MatterportB} \\ 
    \multicolumn{2}{l}{} & \multicolumn{3}{c}{$R$} & \phantom{a} & \multicolumn{3}{c}{${t}$} & \phantom{a}& \phantom{a} & \multicolumn{3}{c}{$R$} & \phantom{a} & \multicolumn{3}{c}{${t}$}                                    \\
    \cmidrule{3-5} \cmidrule{7-9} \cmidrule{12-14} \cmidrule{16-18} 
    \multicolumn{2}{l}{} & mean ($^\circ$) & med ($^\circ$) & rank&& mean ($^\circ$) & med ($^\circ$) & rank &&& mean ($^\circ$) & med ($^\circ$) & rank && mean ($^\circ$) & med ($^\circ$) & rank\\
    \midrule  
    \multirow{4}{*}{DirectionNet }  & \multirow{1}{*}{9D}   & \B{3.96} & 2.28 & \B{2.76} && \B{14.17} & \B{6.46} & \B{3.29}  &&& 13.60 & \B{3.54} & \B{2.89} && \B{21.26} & \B{8.90} & \B{3.44}                \\
    & \multirow{1}{*}{6D}   & 4.30 & \B{2.22} & 2.79 && 16.37 & 7.07 & 3.29 &&& 14.85 & 3.69 & 3.45 && 23.60 & 9.42 & 3.79   \\
    & \multirow{1}{*}{9D-Single}  & 4.55 & 3.11 & 3.83 && 21.65 & 10.53 & 4.71 &&& \B{13.37} & 4.00 & 2.85 && 28.41 & 13.27 & 4.26     \\
    \cmidrule{2-18}
    & \multirow{1}{*}{Quat.}   & 23.32 & 23.00 &8.25 && 39.85 & 24.85 & 6.22 &&& 37.09  & 25.25 & 7.13 && 49.39  & 31.59 & 6.94  \\
    \midrule
    \multirow{4}{*}{Regression}  & \multirow{1}{*}{Bin\&Delta} & 6.93 & 4.71 & 5.28 && 22.84 & 10.16 & 3.73 &&& 31.54 & 22.98 & 6.45 && 29.45  & 14.30 & 5.14   \\
    & \multirow{1}{*}{Spherical} & 10.68 & 7.98 & 6.79 && 40.09 & 22.85 & 6.36   &&& 32.94 & 20.56 & 6.42 && 51.00 & 33.18 & 8.40    \\
    & \multirow{1}{*}{6D} & 5.73 & 3.66 & 3.79 && 35.75 & 21.89 & 6.32 &&& 18.23 & 7.69 & 4.29 && 39.06 & 25.07 & 5.69     \\
    & \multirow{1}{*}{Quat.}  & 15.40 & 12.66 & 6.86 && 41.57 & 21.47 & 7.18 &&& 28.38 & 19.23 & 6.19 && 48.99  & 34.94 &  7.63\\
    \midrule
    \multirow{2}{*}{SIFT}  & \multirow{1}{*}{LMedS}   & 25.55 & 5.63 & 7.71 && 35.53 & 14.84 & 6.20 &&&  36.58 & 10.54 &  8.13 && 42.67 & 26.64 & 6.06   \\
    & \multirow{1}{*}{RANSAC} & 19.33 & 6.66 & 7.31 && 45.04 & 29.78 & 8.08   &&& 31.30 & 9.55 & 7.74 && 47.74  & 26.19 &  6.19    \\
    \bottomrule\\
  \end{tabular}
  \end{adjustbox}
    \caption{\textbf{Quantitative results on the Matterport datasets.} \label{tab:QuantitativeResults}
  We report the mean and median angular error in degrees, as well as average rank of each method over all test pairs. Rotation ($R$) and translation ($t$) shown separately.
  }
\end{table*}

\section{Experiments}
To evaluate our method on challenging data exhibiting a wide range of relative motion, we generate image pairs from existing panoramic image collections. Image pairs are generated by sampling pairs of panoramas from a common scene, then projecting them to overlapping planar perspective views. By varying the camera viewing angles we create a dataset with varied relative poses and overlap. Specific details of each dataset are provided below. Unless specified otherwise, our training image pairs have a resolution of 256$\times$256 and a $90^\circ$ FoV. In all cases, we generate 1M training pairs and $\sim$1K test pairs. \textit{Note that there is no overlap between the train and test scenes}. See appendix \ref{appdataset} for more details.

\medskip
\noindent \textbf{InteriorNet}~\cite{InteriorNet18} is a synthetic dataset with 560 scenes with panoramas rendered along smooth camera trajectories which we sample at random strides. \InteriorNetA is constructed to have rotations up to 30$^\circ$, while \InteriorNetB has varied FoV ($60^\circ$ to $90^\circ$) and rotations up to 40$^\circ$.

\medskip
\noindent \textbf{Matterport3D}~\cite{Matterport3D} contains 10K \emph{real} panoramas captured from locations $\sim$2.25m apart covering 90 scenes. \MatterportA is constructed to have rotations up to 45$^\circ$, and \MatterportB~up to 90$^\circ$. These are more challenging than InterionNet due to a wider baseline and smaller overlap.

\subsection{Training details} \label{ImplementationDetails}
We train with loss weights $\lambda_{\sigma}=0.1$, $\lambda_D=8{\times}10^7$, and Adam~\cite{adam} with a learning rate of 1e-3 and a batch size of 20. The ground truth distributions are generated by the von Mises-Fisher distribution with concentration $\kappa = 10.0$ ($\kappa$ is analogous to $\frac{1}{\sigma^2}$ in a Gaussian distribution).
The derotated images for InteriorNet have a $90^\circ$ FoV, while the derotated Matterport3D images have a $105^\circ$ FoV and an increased resolution of $344{\times344}$ to compensate for the larger rotations.

\subsection{Baselines}
We now introduce the baselines. Full details can be found in appendix \ref{appbaselines}.

\paragraph{DirectionNet variations.}
We consider multiple variants of our full two-stage model with intermediate derotation. \textbf{DirectionNet-9D} projects three direction vectors onto $\SO(3)$ using SVD for the rotation estimation, while \textbf{DirectionNet-6D} uses a partial Gram-Schmidt projection~\cite{zhou19cvpr} (refer to Sec.~\ref{secprocrustes} for details). To understand the importance of derotation, we also consider a single-stage version without derotation (\textbf{DirectionNet-9D-Single}) which estimates four directions from a single DirectionNet module.

\paragraph{Discrete pose representation alternatives.} To evaluate our choice of representation for the discretized pose space, we consider multiple alternatives: \textbf{Bin\&Delta}~\cite{classificationregression2018} is a hybrid model combining a coarse rotation classification (over clustered quaternions) with a refinement regression network.
To understand if decoupling a 3D rotation into a set of direction vectors is necessary, we introduce \textbf{DirectionNet-Quat} which estimates a discrete distribution over the space of unit quaternions at resolution $32^3.$ The spherical decoder is replaced by a 3D volumetric CNN decoder.
\textbf{3D-RCNN}~\cite{rcnn3d} estimates individual (Euler) angles with a discrete distribution over the quantized circle. 3D-RCNN consistently under-performed Bin\&Delta, see appendix \ref{appbaselines}.

\paragraph{Regression baselines.} We evaluated multiple direct pose regression baselines: \textbf{Spherical regression}~\cite{liao2019} uses a novel spherical exponential activation for regression to $n$-spheres; \textbf{6D}~\cite{zhou19cvpr} regresses 6D outputs followed by a Gram-Schmidt projection for rotations; \textbf{Quaternion} regresses a unit quaternion for the rotation. This is what is used in the camera pose regression modules from PoseNet~\cite{kendall15iccv} and~\cite{melekhov17acivs}. The multiple regression baselines share the same image encoder architecture as our DirectionNet.

\paragraph{Parametric probabilistic pose.}
\textbf{vM}~\cite{deepdirectstat2018} estimates individual (Euler) angles by directly regressing the continuous parameters of a von-Mises distribution (location and concentration) on the circle.
We were unable to train vM successfully on all datasets (see appendix \ref{appbaselines}).

\paragraph{Feature-based baselines.} We consider two versions of the classic correspondence-based pipeline, SIFT features~\cite{SIFT} with robust LMedS~\cite{lmeds} (\textbf{SIFT+LMedS}) or with RANSAC~\cite{ransac} (\textbf{SIFT+RANSAC}). We also consider pipelines with learned components such as \textbf{SuperGlue}~\cite{sarlin20superglue} and \textbf{D2-Net}~\cite{Dusmanu2019CVPR}.

Note, the differences in most baselines are in the rotation representation. Unless specified otherwise, the baselines predict a unit-normalized 3D vector for the camera translation direction. 

\paragraph{Evaluation metrics.}  We report geodesic errors for both rotations and translation directions, separately. Additionally, we rank each method on every test pair, reporting the mean rank across examples (1 is the best possible rank, 10 is the worst possible rank).

\begin{figure}[t!]
\centering
\includegraphics[height=2.2in]{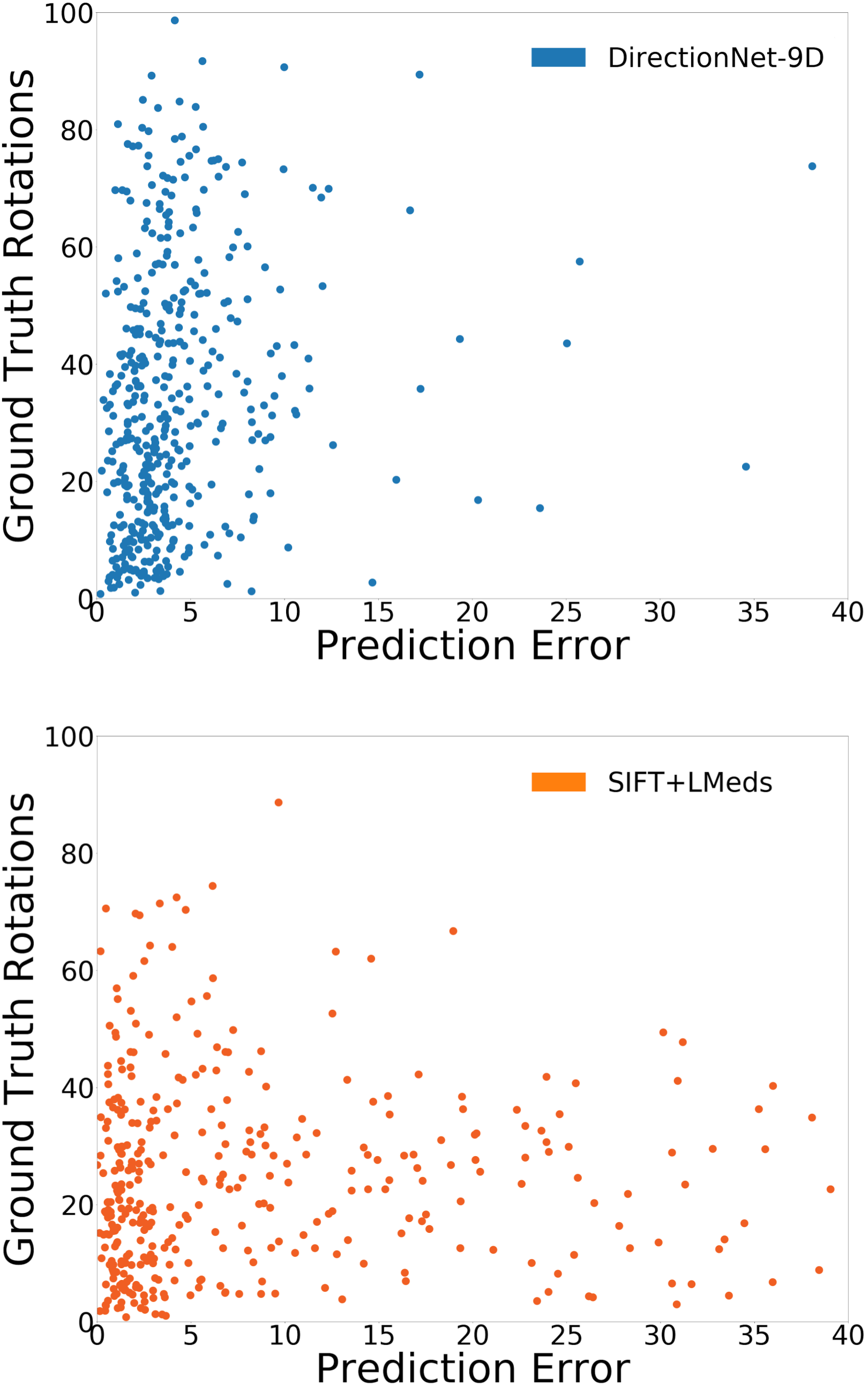}
\hspace{0.4cm}
\includegraphics[height=2.2in]{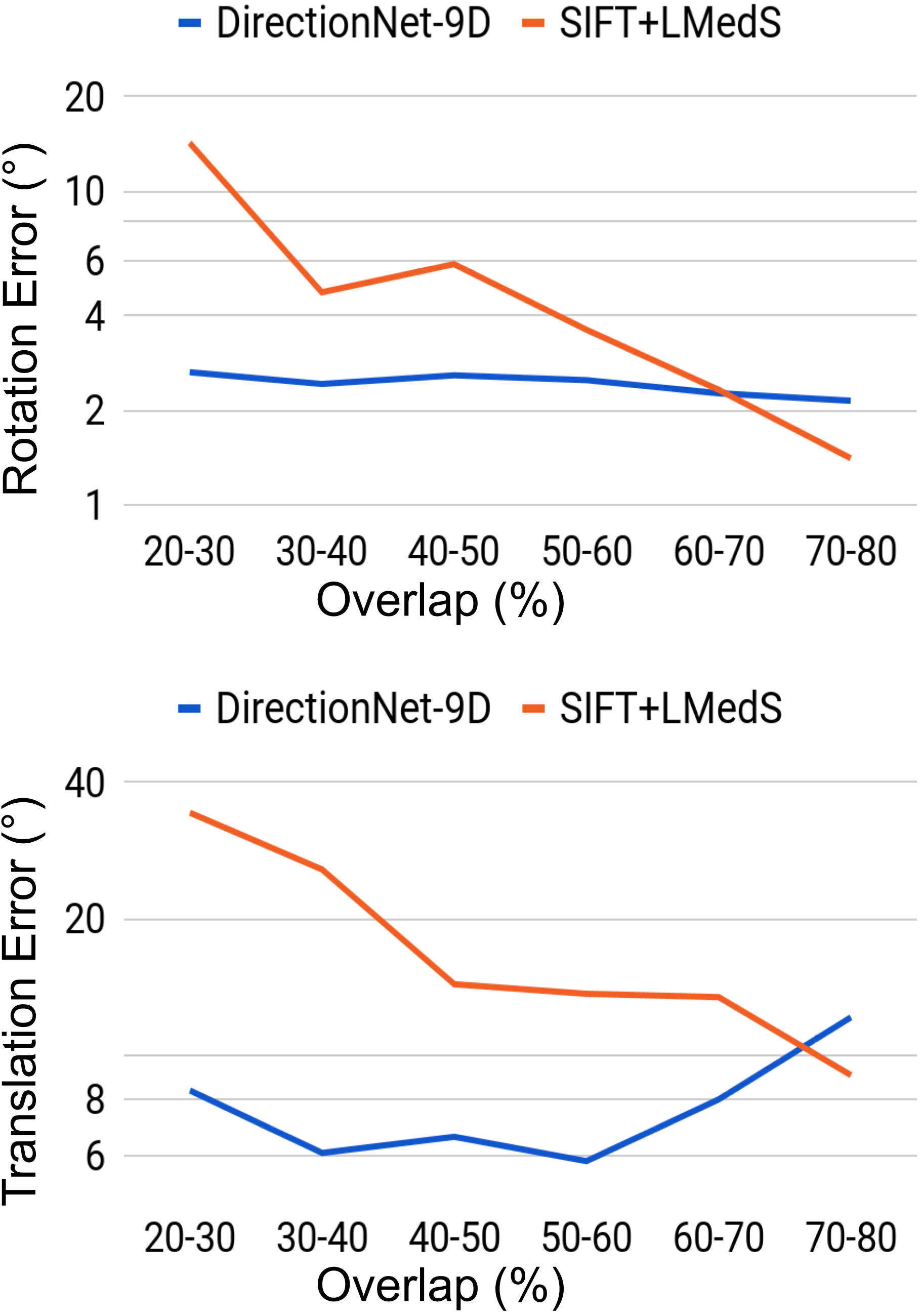}
\caption{
\textbf{(a)} True rotation magnitude ($^\circ$) vs error ($^\circ$). The scatter plot shows that our model is robust in the presence of large relative rotations. \textbf{(b)} Median error ($^\circ$) vs.\ overlap (\%). As image overlap decreases from 90\% to 20\%, the median test errors of our method increases much slower than the SIFT+LMedS. When overlap is very high, local feature based techniques are still superior.
\label{fig:OursVsFeatures}}
\end{figure}

\begin{table}[t!]
   \centering
  \begin{adjustbox}{width=1.\linewidth}
  \begin{tabular}{lccccc}
    \toprule 
    \MatterportA & \multicolumn{2}{c}{$R$} & & \multicolumn{2}{c}{${t}$} \\
    \cmidrule{2-3} \cmidrule{5-6}
    & mean ($^\circ$) & med ($^\circ$) & & mean ($^\circ$) & med ($^\circ$) \\
    \midrule 
    SuperGlue (indoor)\cite{sarlin20superglue}  & 8.34 & 5.22 & & 21.08 & 11.86 \\
    D2-Net~\cite{Dusmanu2019CVPR} & 18.79 & 7.12 & & 41.58 & 25.56 \\
    \midrule
    DirectionNet-9D & \textbf{3.96} & \textbf{2.28} & & \textbf{14.17} & \textbf{6.46} \\
    \bottomrule
    \toprule 
    \MatterportB & \multicolumn{2}{c}{$R$} & & \multicolumn{2}{c}{${t}$} \\
    \cmidrule{2-3} \cmidrule{5-6}
    & mean ($^\circ$) & med ($^\circ$) & & mean ($^\circ$) & med ($^\circ$) \\
    \midrule  
    SuperGlue (indoor)\cite{sarlin20superglue}  & \textbf{13.23} & 7.19 & & 29.90 & 13.28 \\
    D2-Net~\cite{Dusmanu2019CVPR} & 21.03 & 7.74 & & 43.82 & 27.30 \\
    \midrule
    DirectionNet-9D & 13.60 & \textbf{3.54} & & \textbf{21.26} & \textbf{8.90} \\
    \bottomrule\\
  \end{tabular}
  \end{adjustbox}
  \caption{\label{tab:localfeatMatterportA}Performance of learned feature-based methods on the Matterport A and B datasets.}
\end{table}

\subsection{Analysis}\label{sec:analysis}
Table~\ref{tab:QuantitativeResults} reports the quantitative results of the different methods on the Matterport datasets (see Table~\ref{tab:QuantitativeResultsInteriorNet} for results on the InteriorNet datasets). We begin by considering the two main questions for analyzing our approach.

\textit{Can relative camera pose be better learned using a discrete pose distribution versus direct regression?} We observe that our DirectionNet outperforms all baselines on each metric for each dataset. A particularly illustrative comparison is DirectionNet-6D vs 6D regression, where for the same pose representation, prediction via a discrete distribution is consistently better. We do not see the same improvement for DirectionNet-Quat over Quaternion regression, however, which indicates our choice of the lower-dimensional spherical/directional representation is important (see discussion below). Finally, we observe that predicting a parametric probabilistic representation of pose does not help (vM~\cite{deepdirectstat2018} performed 5x worse than DirectionNet, see appendix \ref{appbaselines}).

\textit{Is our directional representation better than alternatives for estimating discrete distributions over relative camera pose?}
Both DirectionNet-6D/9D and DirectionNet-9D-Single consistently outperform alternatives which also use a quantized output space in some way, namely Bin\&Delta~\cite{classificationregression2018}, DirectionNet-Quat, and 3D-RCNN~\cite{rcnn3d}. Our DirectionNet-Quat baseline predicts a distribution over the half hypersphere in $S^3$. Its poor performance supports our hypothesis that the smaller resolution allowed by its $O(N^3)$ space requirements limits performance, whereas our directional models require just $O(N^2)$ space.

\paragraph{Feature-based approaches.} In this work our aim is to understand if relative pose regression can be improved using a discrete pose distribution representation, especially in the difficult wide-baseline setting. Thus, our primary experimental analysis is a comparison of different regression and probabilistic techniques. For completeness, we also evaluate feature-based approaches. Unsurprisingly, the feature-based methods have different performance characteristics compared to the learned direct methods.  For example, for those image pairs where feature extraction and matching are likely successful (e.g.\ high overlap), the estimated motion based on SIFT features is consistently better than any learned technique (Fig.~\ref{fig:OursVsFeatures}-b). We note that in the best cases the SIFT methods both regularly reach sub-$1^\circ$ errors in relative rotation estimation, while all of the learning methods rarely reach errors that low.  However, our dataset construction intentionally includes a large fraction of large-motion pairs likely to have low overlap, and this drives down the overall performance of these methods (Fig.~\ref{fig:OursVsFeatures}-a). In addition to classic feature-based methods, we compare with learned descriptors and matching pipelines. Table~\ref{tab:localfeatMatterportA} shows results using pretrained models for SuperGlue~\cite{sarlin20superglue} and D2-Net~\cite{Dusmanu2019CVPR}.  SuperGlue has slightly better performance than DirectionNet on mean rotation error for Matterport-B, but in general DirectionNet outperforms the learned feature-based methods. This is not surprising since our datasets include many image pairs where keypoint detection and matching can be difficult.

\begin{figure*}[p]
\begin{center}
\includegraphics[width=0.87\linewidth]{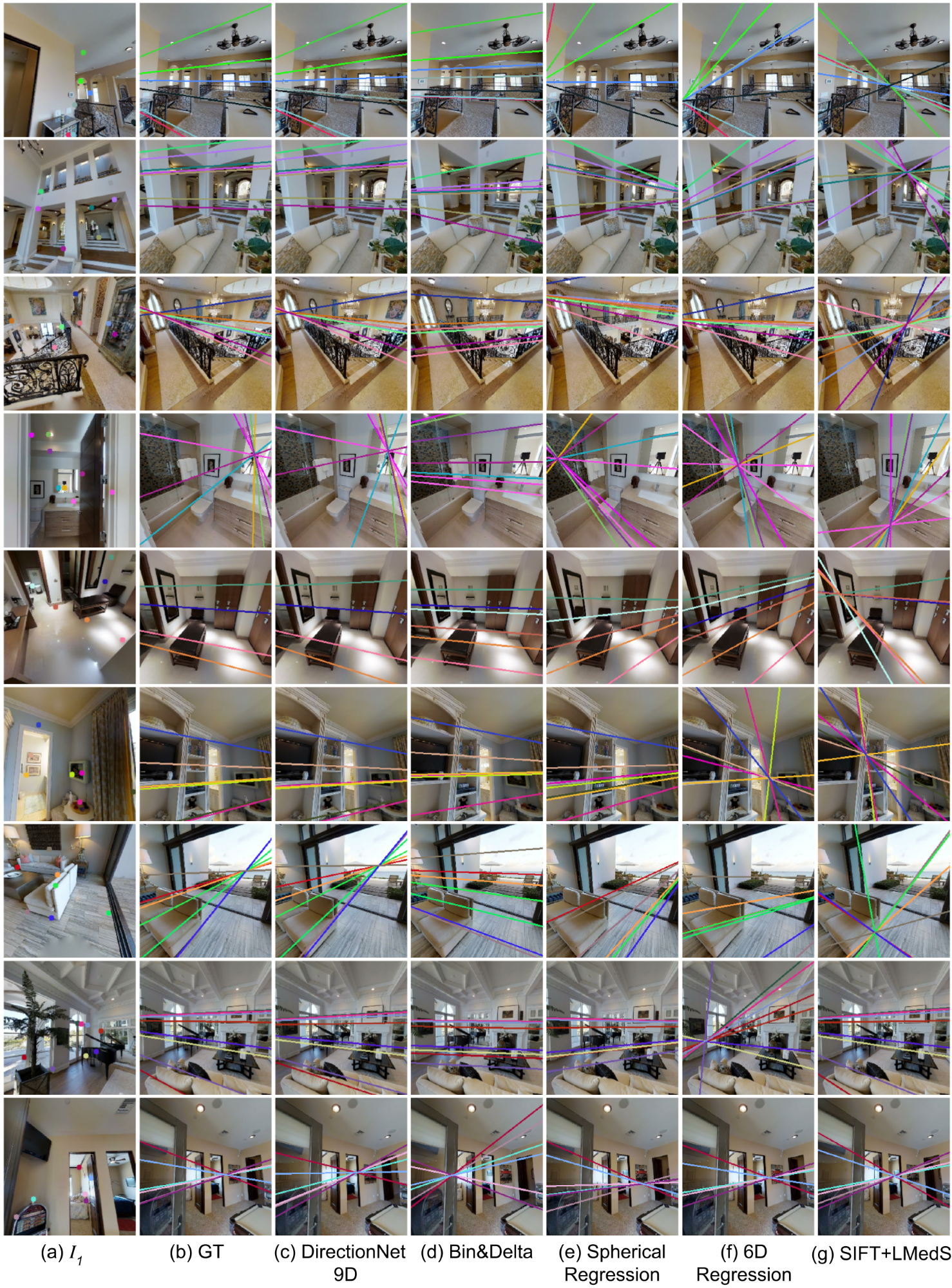}
\end{center}
   \caption{\textbf{Qualitative evaluation on \MatterportB.} 
   Any point in one image plane corresponds to a ray shooting from the optical center, which could be projected to the other image plane as the \textit{epipolar line}. (a) We draw a number of points detected by SIFT in different colors on each target image $I_1$, and (b) show their corresponding epipolar lines on the source image using the ground truth pose, (c) visualizations from our DirectionNet-9D, (d) Bin \& Delta, (e) spherical regression, (f) 6D regression, (g) SIFT+LMedS. Most examples demonstrates some of the most difficult scenarios, such as drastic change in viewpoint and significant occlusion. The last two rows show that SIFT+LMedS can outperform the others in the case of smaller motions for which the the feature-based approach can find reliable feature correspondences.
   } 
\label{fig:QualitativeResults}
\end{figure*}

\paragraph{Qualitative Results.} Figure~\ref{fig:QualitativeResults} shows results on the challenging real Matterport data which includes large baselines and occlusions. We qualitatively assess each method by visualizing epipolar lines after prediction. We see that DirectionNet can still recover the correct relative pose in never-seen test scenes even when presented with extreme motions.

\paragraph{Generalization.} To demonstrate the generalization ability of our model, we train DirectionNet-9D on InteriorNet-A (synthetic) and test it on Matterport-A (real). The mean and the median errors of the rotation are 8.42$^\circ$ and 5.13$^\circ$, and 20.71$^\circ$ and 8.60$^\circ$ for translation. \textit{Even without any fine-tuning on real data our approach still outperforms most baselines which had the benefit of training on Matterport-A.}
To test the model's performance on outdoor scenes, we trained on a subset of KITTI~\cite{Geiger2012CVPR}. DirectionNet gives 9.19$^\circ$ mean rotation error and 19.36$^\circ$ translation error while the best baseline gives 13.44$^\circ$ and 22.53$^\circ$ for rotation and translation respectively. See Table~\ref{tab:kitti}.

\section{Conclusion}
The results presented above tell a consistent story. Models that regress relative pose directly from wide-baseline image pairs can be improved by estimating a discrete probability distribution in the pose space. Our approach effectively executes this idea by operating on a factorized pose space that is lower dimensional than the 5D pose space and suitable for discretized outputs. Evaluated on challenging synthetic and real wide-baseline datasets, DirectionNet generally outperforms regression models, parametric probabilistic models, and alternative discretization schemes.
{\small

\begin{thebibliography}{10}\itemsep=-1pt

\bibitem{tensorflow2015-whitepaper}
Mart\'{\i}n Abadi, Ashish Agarwal, Paul Barham, Eugene Brevdo, Zhifeng Chen,
  Craig Citro, Greg~S. Corrado, Andy Davis, Jeffrey Dean, Matthieu Devin,
  Sanjay Ghemawat, Ian Goodfellow, Andrew Harp, Geoffrey Irving, Michael Isard,
  Yangqing Jia, Rafal Jozefowicz, Lukasz Kaiser, Manjunath Kudlur, Josh
  Levenberg, Dan Man\'{e}, Rajat Monga, Sherry Moore, Derek Murray, Chris Olah,
  Mike Schuster, Jonathon Shlens, Benoit Steiner, Ilya Sutskever, Kunal Talwar,
  Paul Tucker, Vincent Vanhoucke, Vijay Vasudevan, Fernanda Vi\'{e}gas, Oriol
  Vinyals, Pete Warden, Martin Wattenberg, Martin Wicke, Yuan Yu, and Xiaoqiang
  Zheng.
\newblock {TensorFlow}: Large-scale machine learning on heterogeneous systems,
  2015.
\newblock Software available from tensorflow.org.

\bibitem{afsari2011}
Bijan Afsari.
\newblock Riemannian {$L^p$} center of mass: Existence, uniqueness, and
  convexity.
\newblock {\em Proceedings of the American Mathematical Society}, 139, 02 2011.

\bibitem{brachmann17cvpr}
Eric Brachmann, Alexander Krull, Sebastian Nowozin, Jamie Shotton, Frank
  Michel, Stefan Gumhold, and Carsten Rother.
\newblock Dsac - differentiable ransac for camera localization.
\newblock In {\em IEEE Conference on Computer Vision and Pattern Recognition
  (CVPR)}, page~9, 2018.

\bibitem{Campbell_2019_CVPR}
Dylan Campbell, Lars Petersson, Laurent Kneip, Hongdong Li, and Stephen Gould.
\newblock The alignment of the spheres: Globally-optimal spherical mixture
  alignment for camera pose estimation.
\newblock In {\em IEEE Conference on Computer Vision and Pattern Recognition
  (CVPR)}, June 2019.

\bibitem{Matterport3D}
Angel Chang, Angela Dai, Thomas Funkhouser, Maciej Halber, Matthias Niessner,
  Manolis Savva, Shuran Song, Andy Zeng, and Yinda Zhang.
\newblock {Matterport3D}: Learning from {RGB-D} data in indoor environments.
\newblock In {\em International Conference on 3D Vision (3DV)}, 2017.

\bibitem{chen2020survey}
Changhao Chen, Bing Wang, Chris~Xiaoxuan Lu, Niki Trigoni, and Andrew Markham.
\newblock A survey on deep learning for localization and mapping: Towards the
  age of spatial machine intelligence.
\newblock {\em arXiv preprint arXiv:2006.12567}, 2020.

\bibitem{s.2018spherical}
Taco~S. Cohen, Mario Geiger, Jonas Köhler, and Max Welling.
\newblock Spherical {CNN}s.
\newblock In {\em International Conference on Learning Representations (ICLR)},
  2018.

\bibitem{poserbpf}
Xinke Deng, Arsalan Mousavian, Yu Xiang, Fei Xia, Timothy Bretl, and Dieter
  Fox.
\newblock Pose{RBPF}: A {R}ao-{B}lackwellized {P}article {F}ilter for {6D}
  {O}bject {P}ose {E}stimation.
\newblock In {\em Proceedings of Robotics: Science and Systems},
  FreiburgimBreisgau, Germany, June 2019.

\bibitem{detone16rssw}
Daniel DeTone, Tomasz Malisiewicz, and Andrew Rabinovich.
\newblock Deep image homography estimation.
\newblock In {\em RSS Workshop on Limits and Potentials of Deep Learning in
  Robotics}, 2016.

\bibitem{DeTone_2018_CVPR_Workshops}
Daniel DeTone, Tomasz Malisiewicz, and Andrew Rabinovich.
\newblock Superpoint: Self-supervised interest point detection and description.
\newblock In {\em IEEE Conference on Computer Vision and Pattern Recognition
  (CVPR) Workshops}, June 2018.

\bibitem{Dusmanu2019CVPR}
Mihai Dusmanu, Ignacio Rocco, Tomas Pajdla, Marc Pollefeys, Josef Sivic,
  Akihiko Torii, and Torsten Sattler.
\newblock {D2-Net: A Trainable CNN for Joint Detection and Description of Local
  Features}.
\newblock In {\em IEEE Conference on Computer Vision and Pattern Recognition
  (CVPR)}, 2019.

\bibitem{en18eccvw}
Sovann En, Alexis Lechervy, and Frédéric Jurie.
\newblock Rpnet: An end-to-end network for relative camera pose estimation.
\newblock In {\em European Conference on Computer Vision Workshops (ECCVW)},
  2018.

\bibitem{dso}
Jakob Engel, Vladlen Koltun, and Daniel Cremers.
\newblock Direct sparse odometry.
\newblock {\em IEEE Transactions on Pattern Analysis and Machine Intelligence},
  40(3):611--625, 2017.

\bibitem{esteves18eccv}
Carlos Esteves, Christine Allen-Blanchette, Ameesh Makadia, and Kostas
  Daniilidis.
\newblock Learning {SO}(3) equivariant representations with spherical {CNN}s.
\newblock In {\em European Conference on Computer Vision (ECCV)}, September
  2018.

\bibitem{esteves19icml}
Carlos Esteves, Avneesh Sud, Zhengyi Luo, Kostas Daniilidis, and Ameesh
  Makadia.
\newblock Cross-domain {3D} equivariant image embeddings.
\newblock In {\em International Conference on Machine Learning, {ICML}}, 2019.

\bibitem{ransac}
Martin~A. Fischler and Robert~C. Bolles.
\newblock Random sample consensus: A paradigm for model fitting with
  applications to image analysis and automated cartography.
\newblock {\em Commun. ACM}, 24(6):381--395, June 1981.

\bibitem{Fragoso2017PatchCF}
Victor Fragoso, Chunhui Liu, Aayush Bansal, and Deva Ramanan.
\newblock Patch correspondences for interpreting pixel-level cnns.
\newblock {\em arXiv: Computer Vision and Pattern Recognition}, 2017.

\bibitem{Geiger2012CVPR}
Andreas Geiger, Philip Lenz, and Raquel Urtasun.
\newblock Are we ready for autonomous driving? the kitti vision benchmark
  suite.
\newblock In {\em IEEE Conference on Computer Vision and Pattern Recognition
  (CVPR)}, 2012.

\bibitem{Gilitschenski2020Deep}
Igor Gilitschenski, Roshni Sahoo, Wilko Schwarting, Alexander Amini, Sertac
  Karaman, and Daniela Rus.
\newblock Deep orientation uncertainty learning based on a bingham loss.
\newblock In {\em International Conference on Learning Representations}, 2020.

\bibitem{mvg03book}
Richard Hartley and Andrew Zisserman.
\newblock {\em Multiple View Geometry in Computer Vision}.
\newblock Cambridge University Press, New York, NY, USA, 2 edition, 2003.

\bibitem{hartley8pt}
Richard~I. Hartley.
\newblock In defense of the eight-point algorithm.
\newblock {\em IEEE Transactions on Pattern Analysis and Machine Intelligence},
  19(6):580--593, June 1997.

\bibitem{resnet}
Kaiming He, Xiangyu Zhang, Shaoqing Ren, and Jian Sun.
\newblock Identity mappings in deep residual networks.
\newblock In {\em European Conference on Computer Vision (ECCV)}, 2016.

\bibitem{kendall15iccv}
Alex Kendall, Matthew Grimes, and Roberto Cipolla.
\newblock Posenet: A convolutional network for real-time 6-dof camera
  relocalization.
\newblock In {\em IEEE International Conference on Computer Vision (ICCV)},
  page 2938–2946, 2015.

\bibitem{adam}
Diederik~P. {Kingma} and Jimmy {Ba}.
\newblock {Adam: A Method for Stochastic Optimization}.
\newblock In {\em International Conference for Learning Representations}, 2015.

\bibitem{kostelec2008}
Peter~J. Kostelec and Daniel~N. Rockmore.
\newblock Ffts on the rotation group.
\newblock {\em Journal of Fourier Analysis and Applications}, 14(2):145--179,
  Apr 2008.

\bibitem{rcnn3d}
A. {Kundu}, Y. {Li}, and J.~M. {Rehg}.
\newblock {3D-RCNN}: Instance-level {3D} object reconstruction via
  render-and-compare.
\newblock In {\em IEEE Conference on Computer Vision and Pattern Recognition
  (CVPR)}, pages 3559--3568, 2018.

\bibitem{levinson20neurips}
Jake Levinson, Carlos Esteves, Kefan Chen, Noah Snavely, Angjoo Kanazawa,
  Afshin Rostamizadeh, and Ameesh Makadia.
\newblock An analysis of {SVD} for deep rotation estimation.
\newblock In {\em Advances in Neural Information Processing Systems 34}, 2020.

\bibitem{InteriorNet18}
Wenbin Li, Sajad Saeedi, John McCormac, Ronald Clark, Dimos Tzoumanikas, Qing
  Ye, Yuzhong Huang, Rui Tang, and Stefan Leutenegger.
\newblock Interiornet: Mega-scale multi-sensor photo-realistic indoor scenes
  dataset.
\newblock In {\em British Machine Vision Conference (BMVC)}, 2018.

\bibitem{li18eccv}
Yi Li, Gu Wang, Xiangyang Ji, Yu Xiang, and Dieter Fox.
\newblock Deepim: Deep iterative matching for 6d pose estimation.
\newblock In {\em European Conference on Computer Vision (ECCV)}, September
  2018.

\bibitem{liao2019}
Shuai Liao, Efstratios Gavves, and Cees G.~M. Snoek.
\newblock Spherical regression: Learning viewpoints, surface normals and {3D}
  rotations on $n$-spheres.
\newblock In {\em IEEE Conference on Computer Vision and Pattern Recognition
  (CVPR)}, 2019.

\bibitem{LonguetHiggins81}
H.~C. Longuet-Higgins.
\newblock A computer algorithm for reconstructing a scene from two projections.
\newblock {\em Nature}, 293(5828):133--135, 1981.

\bibitem{SIFT}
David~G. Lowe.
\newblock Distinctive image features from scale-invariant keypoints.
\newblock {\em International Journal of Computer Vision}, 60(2):91--110, 2004.

\bibitem{luvizon19humanpose}
Diogo~C. Luvizon, Hedi Tabia, and David Picard.
\newblock Human pose regression by combining indirect part detection and
  contextual information.
\newblock {\em Computers \& Graphics}, 85:15--22, Dec 2019.

\bibitem{mahendran17cvprw}
S. Mahendran, H. Ali, and R. Vidal.
\newblock 3d pose regression using convolutional neural networks.
\newblock In {\em IEEE Conference on Computer Vision and Pattern Recognition
  Workshops (CVPRW)}, 2017.

\bibitem{classificationregression2018}
Siddharth Mahendran, Haider Ali, and Ren{\'e} Vidal.
\newblock A mixed classification-regression framework for {3D} pose estimation
  from 2d images.
\newblock {\em The British Machine Vision Conference (BMVC)}, 2018.

\bibitem{makadia07ijcv}
Ameesh Makadia, Christopher Geyer, and Kostas Daniilidis.
\newblock Correspondence-free structure from motion.
\newblock {\em International Journal of Computer Vision}, 75(3):311--327, 2007.

\bibitem{manhardt2018}
Fabian Manhardt, Diego~Martin Arroyo, Christian Rupprecht, Benjamin Busam,
  Tolga Birdal, Nassir Navab, and Federico Tombari.
\newblock Explaining the {A}mbiguity of {O}bject {D}etection and {6D} {P}ose
  {F}rom {V}isual {D}ata.
\newblock In {\em Proceedings of the IEEE/CVF International Conference on
  Computer Vision (ICCV)}, October 2019.

\bibitem{mardia1975}
K.~V. Mardia.
\newblock Statistics of directional data.
\newblock {\em Journal of the Royal Statistical Society. Series B
  (Methodological)}, 37(3):349--393, 1975.

\bibitem{melekhov17acivs}
Iaroslav Melekhov, Juha Ylioinas, Juho Kannala, and Esa Rahtu.
\newblock Relative camera pose estimation using convolutional neural networks.
\newblock In {\em International Conference on Advanced Concepts for Intelligent
  Vision Systems}, 2017.

\bibitem{MousavianCVPR17}
Arsalan Mousavian, Dragomir Anguelov, John Flynn, and Jana Kosecka.
\newblock 3d bounding box estimation using deep learning and geometry.
\newblock In {\em IEEE Conference on Computer Vision and Pattern Recognition
  (CVPR)}, 2017.

\bibitem{newell2016stacked}
Alejandro Newell, Kaiyu Yang, and Jia Deng.
\newblock Stacked hourglass networks for human pose estimation.
\newblock In {\em European Conference on Computer Vision (ECCV)}, pages
  483--499. Springer, 2016.

\bibitem{nguyen18iral}
Ty Nguyen, Steven~W. Chen, Shreyas~S. Shivakumar, Camillo~J. Taylor, and Vijay
  Kumar.
\newblock Unsupervised deep homography: A fast and robust homography estimation
  model.
\newblock In {\em IEEE Robotics and Automation Letters}, volume~3, pages
  2346--2353, 2018.

\bibitem{nister03cvpr}
David Nister.
\newblock An efficient solution to the five-point relative pose problem.
\newblock In {\em IEEE Conference on Computer Vision and Pattern Recognition
  (CVPR)}, pages 195--202, 2003.

\bibitem{fivepoint}
D. {Nister}.
\newblock An efficient solution to the five-point relative pose problem.
\newblock {\em IEEE Transactions on Pattern Analysis and Machine Intelligence},
  26(6):756--770, 2004.

\bibitem{lfnet}
Yuki Ono, Eduard Trulls, Pascal Fua, and Kwang~Moo Yi.
\newblock Lf-net: Learning local features from images.
\newblock In {\em Advances in Neural Information Processing Systems},
  volume~31, 2018.

\bibitem{peretroukhin20rss}
Valentin Peretroukhin, Matthew Giamou, David~M. Rosen, W.~Nicholas Greene,
  Nicholas Roy, and Jonathan Kelly.
\newblock A {S}mooth {R}epresentation of {SO(3)} for {D}eep {R}otation
  {L}earning with {U}ncertainty.
\newblock In {\em Proceedings of {R}obotics: {S}cience and {S}ystems
  {(RSS'20)}}, 2020.

\bibitem{hydranet}
Valentin Peretroukhin, Brandon Wagstaff, and and Jonathan~Kelly.
\newblock Deep {P}robabilistic {R}egression of {E}lements of {SO(3)} using
  {Q}uaternion {A}veraging and {U}ncertainty {I}njection.
\newblock In {\em Proceedings of the IEEE/CVF Conference on Computer Vision and
  Pattern Recognition (CVPR) Workshops}, June 2019.

\bibitem{poursaeed2018eccv}
Omid Poursaeed, Guandao Yang, Aditya Prakash, Qiuren Fang, Hanqing Jiang,
  Bharath Hariharan, and Serge Belongie.
\newblock Deep fundamental matrix estimation without correspondences.
\newblock In {\em European Conference on Computer Vision (ECCV)}, pages
  485--497, 2018.

\bibitem{probst2019cvpr}
Thomas Probst, Danda~Pani Paudel, Ajad Chhatkuli, and Luc~Van Gool.
\newblock Unsupervised learning of consensus maximization for {3D} vision
  problems.
\newblock In {\em IEEE Conference on Computer Vision and Pattern Recognition
  (CVPR)}, June 2019.

\bibitem{deepdirectstat2018}
Sergey Prokudin, Peter Gehler, and Sebastian Nowozin.
\newblock Deep directional statistics: Pose estimation with uncertainty
  quantification.
\newblock In {\em European Conference on Computer Vision (ECCV)}, Sept. 2018.

\bibitem{ransacsurvey}
Rahul Raguram, Jan-Michael Frahm, and Marc Pollefeys.
\newblock A comparative analysis of ransac techniques leading to adaptive
  real-time random sample consensus.
\newblock In {\em European Conference on Computer Vision (ECCV)}, pages
  500--513, 2008.

\bibitem{ranftl2018eccv}
Ren\'e Ranftl and Vladlen Koltun.
\newblock Deep fundamental matrix estimation.
\newblock In {\em European Conference on Computer Vision (ECCV)}, 2018.

\bibitem{RonnebergerFB15}
Olaf Ronneberger, Philipp Fischer, and Thomas Brox.
\newblock {U-Net}: Convolutional networks for biomedical image segmentation.
\newblock In {\em International Conference on Medical Image Computing and
  Computer Assisted Intervention}, 2015.

\bibitem{sarlin20superglue}
Paul-Edouard Sarlin, Daniel DeTone, Tomasz Malisiewicz, and Andrew Rabinovich.
\newblock {SuperGlue}: Learning feature matching with graph neural networks.
\newblock In {\em IEEE Conference on Computer Vision and Pattern Recognition
  (CVPR)}, 2020.

\bibitem{procrustes}
P.H. Schönemann.
\newblock A generalized solution of the orthogonal procrustes problem.
\newblock {\em Psychometrika}, 31:1--10, 1966.

\bibitem{su15iccv}
Hao Su, Charles~R. Qi, Yangyan Li, and Leonidas~J. Guibas.
\newblock Render for cnn: Viewpoint estimation in images using cnns trained
  with rendered {3D} model views.
\newblock In {\em IEEE International Conference on Computer Vision (ICCV)},
  December 2015.

\bibitem{sun2017integral}
Xiao Sun, Bin Xiao, Fangyin Wei, Shuang Liang, and Yichen Wei.
\newblock Integral human pose regression.
\newblock In {\em European Conference on Computer Vision (ECCV)}, 2018.

\bibitem{sundermeyer2018eccv}
Martin Sundermeyer, Zoltan-Csaba Marton, Maximilian Durner, Manuel Brucker, and
  Rudolph Triebel.
\newblock Implicit {3D} orientation learning for 6d object detection from rgb
  images.
\newblock In {\em European Conference on Computer Vision (ECCV)}, September
  2018.

\bibitem{suwajanakorn2018discovery}
Supasorn Suwajanakorn, Noah Snavely, Jonathan~J Tompson, and Mohammad Norouzi.
\newblock Discovery of latent {3D} keypoints via end-to-end geometric
  reasoning.
\newblock In {\em Advances in Neural Information Processing Systems (NeurIPS)},
  pages 2063--2074, 2018.

\bibitem{vpsKpsTulsianiM15}
Shubham Tulsiani and Jitendra Malik.
\newblock Viewpoints and keypoints.
\newblock In {\em IEEE Conference on Computer Vision and Pattern Recognition
  (CVPR)}, 2015.

\bibitem{ummenhoferZUMID16}
Benjamin Ummenhofer, Huizhong Zhou, Jonas Uhrig, Nikolaus Mayer, Eddy Ilg,
  Alexey Dosovitskiy, and Thomas Brox.
\newblock Demon: Depth and motion network for learning monocular stereo.
\newblock In {\em IEEE Conference on Computer Vision and Pattern Recognition
  (CVPR)}, 2017.

\bibitem{deepvo}
Sen Wang, Ronald Clark, Hongkai Wen, and Niki Trigoni.
\newblock Deepvo: Towards end-to-end visual odometry with deep recurrent
  convolutional neural networks.
\newblock In {\em 2017 IEEE International Conference on Robotics and Automation
  (ICRA)}, pages 2043--2050. IEEE, 2017.

\bibitem{d3vo}
Nan Yang, Lukas~von Stumberg, Rui Wang, and Daniel Cremers.
\newblock D3vo: Deep depth, deep pose and deep uncertainty for monocular visual
  odometry.
\newblock In {\em IEEE Conference on Computer Vision and Pattern Recognition
  (CVPR)}, pages 1281--1292, 2020.

\bibitem{yi18cvpr}
Kwang~Moo Yi, Eduard Trulls~Fortuny, Yuki Ono, Vincent Lepetit, Mathieu
  Salzmann, and Pascal Fua.
\newblock Learning to find good correspondences.
\newblock In {\em IEEE Conference on Computer Vision and Pattern Recognition
  (CVPR)}, page~9, 2018.

\bibitem{geonet}
Zhichao Yin and Jianping Shi.
\newblock Geonet: Unsupervised learning of dense depth, optical flow and camera
  pose.
\newblock In {\em IEEE Conference on Computer Vision and Pattern Recognition
  (CVPR)}, pages 1983--1992, 2018.

\bibitem{lmeds}
Zhengyou Zhang, Rachid Deriche, Olivier Faugeras, and Quang-Tuan Luong.
\newblock A robust technique for matching two uncalibrated images through the
  recovery of the unknown epipolar geometry.
\newblock {\em Artificial Intelligence}, pages 87--119, 1995.

\bibitem{zhou2017unsupervised}
Tinghui Zhou, Matthew Brown, Noah Snavely, and David~G. Lowe.
\newblock Unsupervised learning of depth and ego-motion from video.
\newblock In {\em IEEE Conference on Computer Vision and Pattern Recognition
  (CVPR)}, 2017.

\bibitem{zhou19cvpr}
Yi Zhou, Connelly Barnes, Jingwan Lu, Jimei Yang, and Hao Li.
\newblock On the continuity of rotation representations in neural networks.
\newblock In {\em IEEE Conference on Computer Vision and Pattern Recognition
  (CVPR)}, June 2019.

\end{thebibliography}

}
\clearpage
\appendix
\part*{Appendix}
\section{Spherical Padding}\label{apppadding}
\begin{figure}[h]
\centering
\includegraphics[width=0.15\textwidth]{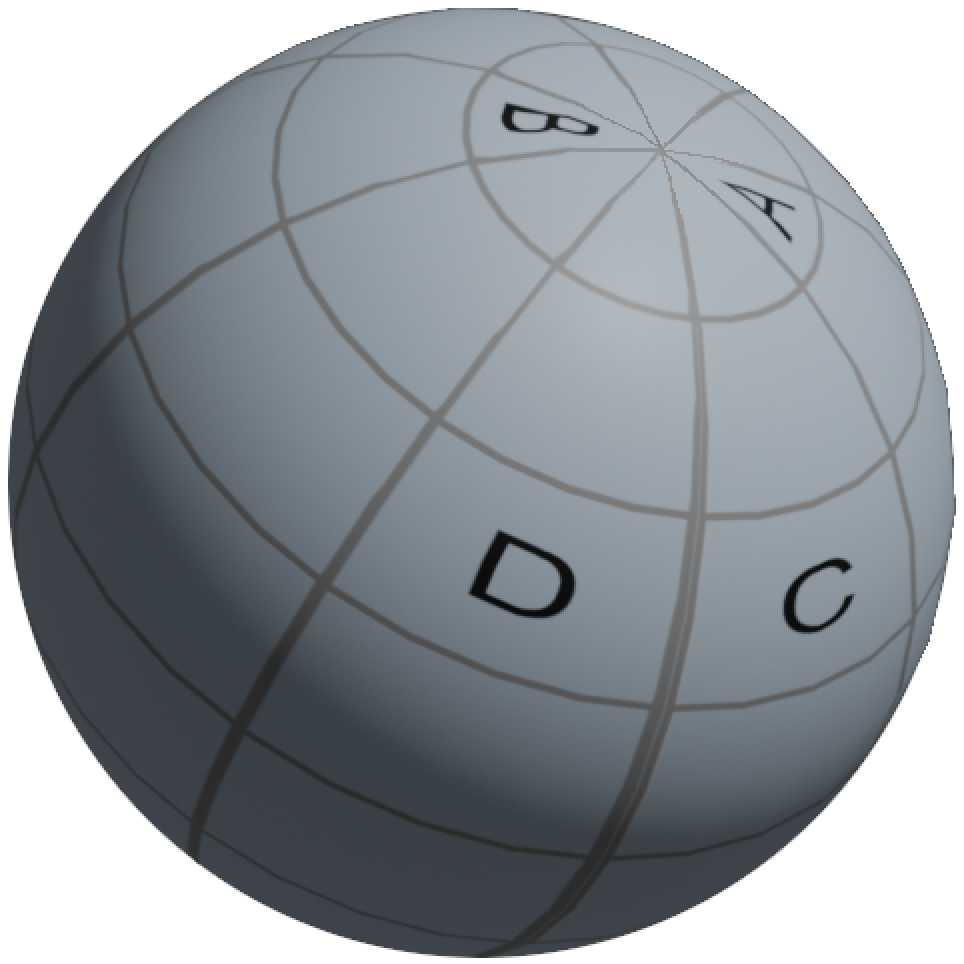}
\hspace{13pt}
\includegraphics[width=0.15\textwidth]{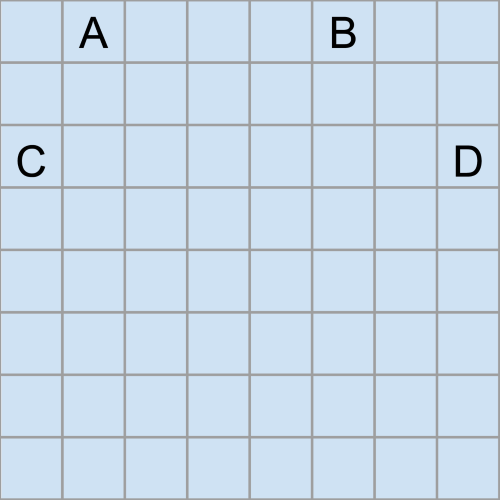}
\caption{\footnotesize Discrete distributions on the sphere (left) are represented internally as equirectangular grids (right).
Although pixels A and B are adjacent on the sphere, as are C and D, they are not adjacent in the  grid. Our spherical padding (shown in Fig.\ 2b, page 4) corrects for this.}
\label{fig:spherepad}
\end{figure}

We propose using spherical padding in our decoder network to reflect the correct topology on a spherical representation (See Figure~\ref{fig:spherepad} for our motivation). 

\section{Dataset Generation}\label{appdataset} 
\begin{figure}[t]
\begin{center}
\subfigure[Data generation]{\label{fig:DataPipeline}\includegraphics[width=0.47\textwidth]{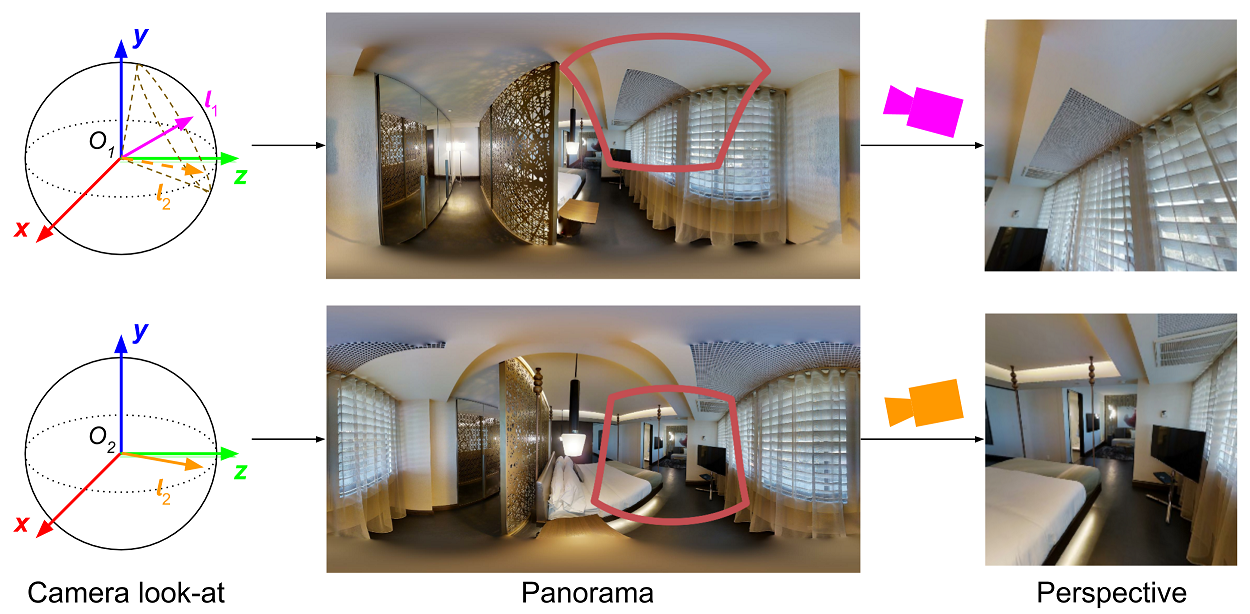}}
\hfill
\subfigure[Epipolar geometry]{\label{fig:EpipolarLine}\includegraphics[width=0.47\textwidth]{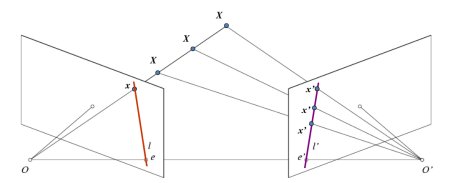}}
\end{center}
  \caption{(a) We randomly sample perspective images from pairs of panoramas by picking the look-at directions $l_1$ and $l_2$ of the source and target cameras based on a heuristic (left). The red boundaries overlaid on the spherical images (center) match the output perspective views (right). (b) Any point $x$ in one image plane corresponds to a ray shooting from its optical center $o$, which represents all possible 3D locations of $x$ in the world. The projection of this ray into the second image plane forms a line called the \textit{epipolar line}, shown in purple in the figure. The 2D point in the second image corresponding to $x$ must lie on this line.
  }
\end{figure}

Since large-scale wide-baseline stereo datasets are difficult to acquire, we create our datasets from corpora of panoramic scene captures by taking pairs of panoramas that observe overlapping parts of scenes, sampling camera look-at directions for each panorama using a heuristic that ensures image overlap, and projecting the panoramas to perspective views with a given field of view. Figure~\ref{fig:DataPipeline} illustrates this process. The look-at direction, $\ell_1$, for the first camera is uniformly sampled from a band around the equator, which is bounded in colatitude \straighttheta $\in$ [$-\frac{\pi}{4}$, $\frac{\pi}{4}$] and azimuth angle \straightphi $\in$ [0, 2\textpi). The look-at direction, $\ell_2$, for the second camera is uniformly sampled from a circular cone centered at the direction $\ell_1$ so that the magnitude of rotations are uniformly distributed in each of our datasets. The limit in latitude prevents the cameras from looking only at the ceiling or floor, which are relatively textureless for many scenes. The aperture of the cone can be adjusted to vary the amount of overlap between image pairs while maintaining variability in the relative camera orientations. Each camera rotation matrix is then constructed from the appropriate look-at vector and the world up vector.

In Figure~\ref{fig:OursVsFeatures}(b)
in the main paper, we show results on the \MatterportB test set grouped by overlap percentage between the input images. Matterport3D panoramas contain depth channels which allows us to calculate the overlap percentage between the input image pair as
\begin{equation}
    O(I_0, I_1) = \min\left(\frac{\lvert I_0\cap I_1\rvert}{\lvert I_0\rvert}, \frac{\lvert I_0\cap I_1\rvert}{\lvert I_1\rvert}\right)
\end{equation}

\section{Training details}\label{apptrainingdetails}
We implemented our model in Tensorflow~\cite{tensorflow2015-whitepaper}. The model was trained asynchronously on 40 Tesla P100 GPUs. A single DirectionNet has approximately 9M parameters. Our full model DirectionNet-9D/6D, which consists of two DirectionNets, contains in total 18M trainable parameters. Each net was trained for 3M steps.

\textbf{Rotation Perturbation.} 
To improve the robustness of DirectionNet-T to rotation estimation errors, we apply data augmentation to its input by perturbing the rotations used for derotation. Given $R \in \textit{SO}(3)$, we perturb it by randomly sampling three unit vectors no further than 15$^\circ$ away from the component vectors of $R$ and projecting the result back onto $SO(3)$. We perturb the estimated rotation from DirectionNet-R before derotating the input images for DirectionNet-T. This perturbation is critical to performance. Without it, the translation error is 4$^\circ$ worse on InteriorNet, and much worse when the rotation range is large (MatterportB).

\section{Relative Pose Baselines}\label{appbaselines}
We now provide additional details of the baselines including the ones not in the main paper.

\begin{itemize}
  \item \textbf{DirectionNet-Quat} directly generates a probability distribution over the half-hypersphere in $S^3$. In this case, the spherical decoder consists of 3D upsampling and 3D convolutional layers. Since the output is on a hypersphere, the discretization requires much higher resolution ($O(N^3)$) compared with our model ($O(N^2)$). DirectionNet-Quat generates output at $32^3$. We believe the limited resolution is partly responsible for the poor performance compared to DirectionNet-9D and -6D.
  
  \item \textbf{Bin\&Delta}~\cite{classificationregression2018} adopts the Bin-Delta hybrid model that consists of a classification network which gives a coarse estimation of the rotation and a regression network that refines the estimate. The rotation space is discretized K-Means clustering on the training data (we use $K=200$). We use the same encoder as ours and DirectionNet-T for the translation.
  
  \item \textbf{Spherical regression}~\cite{liao2019} uses a novel spherical exponential activation on the $n$-sphere to improve the stability of gradients during training. The final outputs of the model are the absolute values of the coordinates of a unit vector in $\mathbb{R}^n$, along with $n$ classification outputs for their signs. We use the same encoder as ours followed by separate two-layer prediction networks (one for a quaternion representation of rotation and one for translation).
  
  \item \textbf{6D} regression uses the same image encoder as ours, followed by two fully connected layers with leaky ReLU and dropout, to produce a 6D continuous representation for the rotation and 3D for the translation. The 6D output is mapped to a rotation matrix with a partial Gram-Schmidt procedure; see \cite{zhou19cvpr} for details. This approach uses a continuous representation for 3D rotation, and consequently facilitates training.
  
  \item The \textbf{quaternion} regression baseline is implemented using same Siamese network as described in \cite{melekhov17acivs} without the spatial pyramid pooling layer, followed by fully connected layers to produce a 4D quaternion and a 3D translation. We normalize the quaternion and the translation during training, and use the same loss as suggested in the paper (L2). In our experiment, we weight the quaternion loss with $\beta=10$, as in the original paper.
  
  \item \textbf{SIFT+LMedS} is a classic technique for recovering the essential matrix from correspondences. Local features are detected in images with SIFT, and subsequently matched across images. These feature matches are filtered with Lowe's proposed distance ratio test. Given the remaining putative correspondences, least median of squares (LMedS) is used to robustly estimate the essential matrix, from which we can recover the rotation and normalized translation direction. We use the OpenCV implementation for all of these steps.
  
  \item \textbf{SIFT+RANSAC} is the same as SIFT+LMedS, with RANSAC instead of LMedS.
  
  \item \textbf{SuperGlue}~\cite{sarlin20superglue} uses CNN and graph neural network to extract and match local features from images. \textbf{D2-Net}~\cite{Dusmanu2019CVPR} trains a single CNN as a dense feature descriptor and a feature detector. Both methods require correspondence/depth supervision from real data, which is not available in our Matterport datasets. We ran pretrained indoor SuperGlue (training code not available) and D2-Net with RANSAC. 

\end{itemize}
\paragraph{Additional baselines.} The following baselines are not presented in the main paper due to the limit of space. For reference, the mean rotation error of our DirectionNet-9D tested on \MatterportA is $3.96^\circ$, on \MatterportB is $13.60^\circ$.  
\begin{itemize}
    \item \textbf{vM}~\cite{deepdirectstat2018} provides a probabilistic formulation for the 2D pose by estimating parameters of von Mises distribution on a circle ($S^1$). We adapt this method to estimate the 3D rotation by producing three von Mises distributions representing the Euler angles. However, the training is hindered by the singularities known in the Euler angles representation\cite{zhou19cvpr}. The mean rotation error tested on \MatterportA is $20.28^\circ$ (5x worse than DirectionNet-9D) and the training diverges on \MatterportB.
    
    \item 3D-RCNN\cite{rcnn3d} uses a classification-regression hybrid model for 2D pose estimation by uniformly discretize the 2D circle into bins. This can be directly adapted to 3D rotation by estimating the three Euler angles. Due to the discontinuity of the Euler angles representation\cite{zhou19cvpr}, the performance is poor compared with the similar hybrid \textbf{Bin\&Delta}  model. The mean rotation error tested on \MatterportA is $18.61^\circ$ and the error on \MatterportB is $32.33^\circ$.
    
    \item \cite{hydranet} combines probabilistic regression and an ensemble of the \textbf{quaternion} regression uisng a multi-headed network called HydraNet. The mean rotation error tested on \MatterportA is $9.38^\circ$ and the error on \MatterportB is $16.09^\circ$.
    
    \item PoseNet\cite{kendall15iccv} relocalizes images in known scenes; we consider relative pose in scenes never seen during training. PoseNet regresses to a 3D position and quaternion, and this is similar to the \textbf{quaternion} regression baseline.

\end{itemize}

\begin{table*}[t!]
  \centering
  \newcommand{\B}[1]{\textbf{#1}}
  \begin{adjustbox}{width=1.\linewidth}
  \begin{tabular}{llcccccccccccccccc}
    
    \toprule 
    \multicolumn{2}{l}{} & \multicolumn{7}{c}{\InteriorNetA} & \phantom{a}& \phantom{a} & \multicolumn{7}{c}{\InteriorNetB} \\
    \multicolumn{2}{l}{} & \multicolumn{3}{c}{$R$} & \phantom{a} & \multicolumn{3}{c}{${t}$} & \phantom{a}& \phantom{a} & \multicolumn{3}{c}{$R$} & \phantom{a} & \multicolumn{3}{c}{${t}$}                                    \\
    \cmidrule{3-5} \cmidrule{7-9} \cmidrule{12-14} \cmidrule{16-18} 
    \multicolumn{2}{l}{} & mean ($^\circ$) & med ($^\circ$) & rank&& mean ($^\circ$) & med ($^\circ$) & rank &&& mean ($^\circ$) & med ($^\circ$) & rank && mean ($^\circ$) & med ($^\circ$) & rank\\
    \midrule   
    \multirow{4}{*}{DirectionNet }  & \multirow{1}{*}{9D}   & \B{2.87} & \B{1.53} & \B{2.30} && \B{12.36} & \B{7.40} & \B{2.59}   &&& 3.88 & \B{2.20} & \B{2.42} && \B{16.36} & \B{9.72} & \B{2.60}               \\
    & \multirow{1}{*}{6D}   & 2.90 & 1.68 & 2.36 && 12.48 & 7.53 & 2.90 &&& \B{3.81} & 2.25 & 2.44 && 16.67 & 10.05 & 2.77\\
    & \multirow{1}{*}{9D-Single}  & 3.93 & 2.61  & 3.29     && 18.17 & 12.56  & 4.71 &&& 4.84 & 3.24 & 3.46 && 26.56 & 19.51 & 4.85   \\
    \cmidrule{2-18}
    & \multirow{1}{*}{Quat.}  & 23.88 & 24.53 & 8.54 && 32.85 & 26.92 & 6.16 &&& 22.11 & 22.43 & 8.55 && 39.45 & 32.05 & 5.50\\
    \midrule
    \multirow{4}{*}{Regression} & \multirow{1}{*}{Bin\&Delta} & 8.79 & 6.59 & 6.10 && 19.53 & 13.45 & 3.87  &&& 6.73 & 4.57 & 5.21 && 31.87 & 22.61 & 4.16   \\
    & \multirow{1}{*}{Spherical} & 19.76 & 15.52 & 8.21 && 31.17 & 23.47 & 5.97 &&& 11.36 & 8.82 & 6.25 && 44.20 & 35.34 & 6.23\\
    & \multirow{1}{*}{6D} & 4.86 & 3.33 & 3.77 && 30.94 & 22.66 & 5.95   &&& 6.03 & 3.95 & 3.99 && 41.29 & 34.41 & 6.04    \\
    & \multirow{1}{*}{Quat.} & 11.14 & 9.64 & 6.27 && 33.14 & 26.23 & 6.31  &&& 13.94 & 11.64 & 6.70 && 45.44 & 38.96 & 6.39   \\
    \midrule
    \multirow{2}{*}{SIFT}  & \multirow{1}{*}{LMedS} & 29.55 & 7.01 & 7.76 && 37.91 & 19.25 & 8.28  &&& 30.46 & 7.64 & 7.92 && 41.83 & 24.50 & 5.58     \\
    & \multirow{1}{*}{RANSAC} & 16.69 & 8.21  & 7.49    && 45.51 & 30.12  & 8.85  &&& 18.75 & 10.52 & 7.10 && 52.46 & 43.56 & 8.85    \\
    \bottomrule
  \end{tabular}
  \end{adjustbox}
    \caption{\textbf{Quantitative results on the InteriorNet datasets.}
  We report the mean and median angular error in degrees, as well as mean rank of each method over all test pairs. Rotation ($R$) and translation ($t$) shown separately.\label{tab:QuantitativeResultsInteriorNet}
  }
\end{table*}

\section{Additional Results and Discussion}\label{appresultsdiscuss}
\begin{figure*}
\centering
\subfigure[\InteriorNetA]{\includegraphics[width=0.22\textwidth]{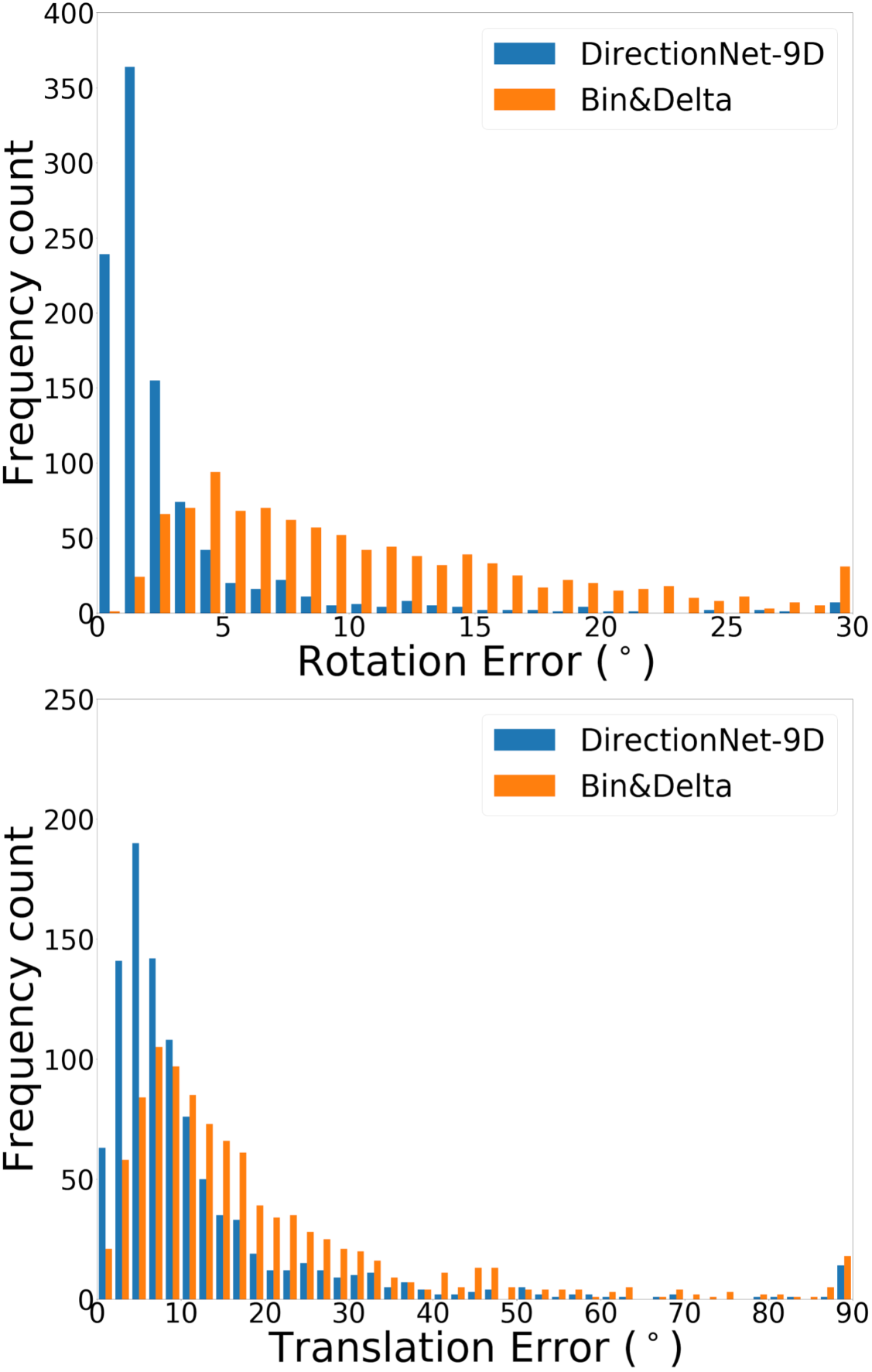}}
\hfill
\subfigure[\InteriorNetB]{\includegraphics[width=0.22\textwidth]{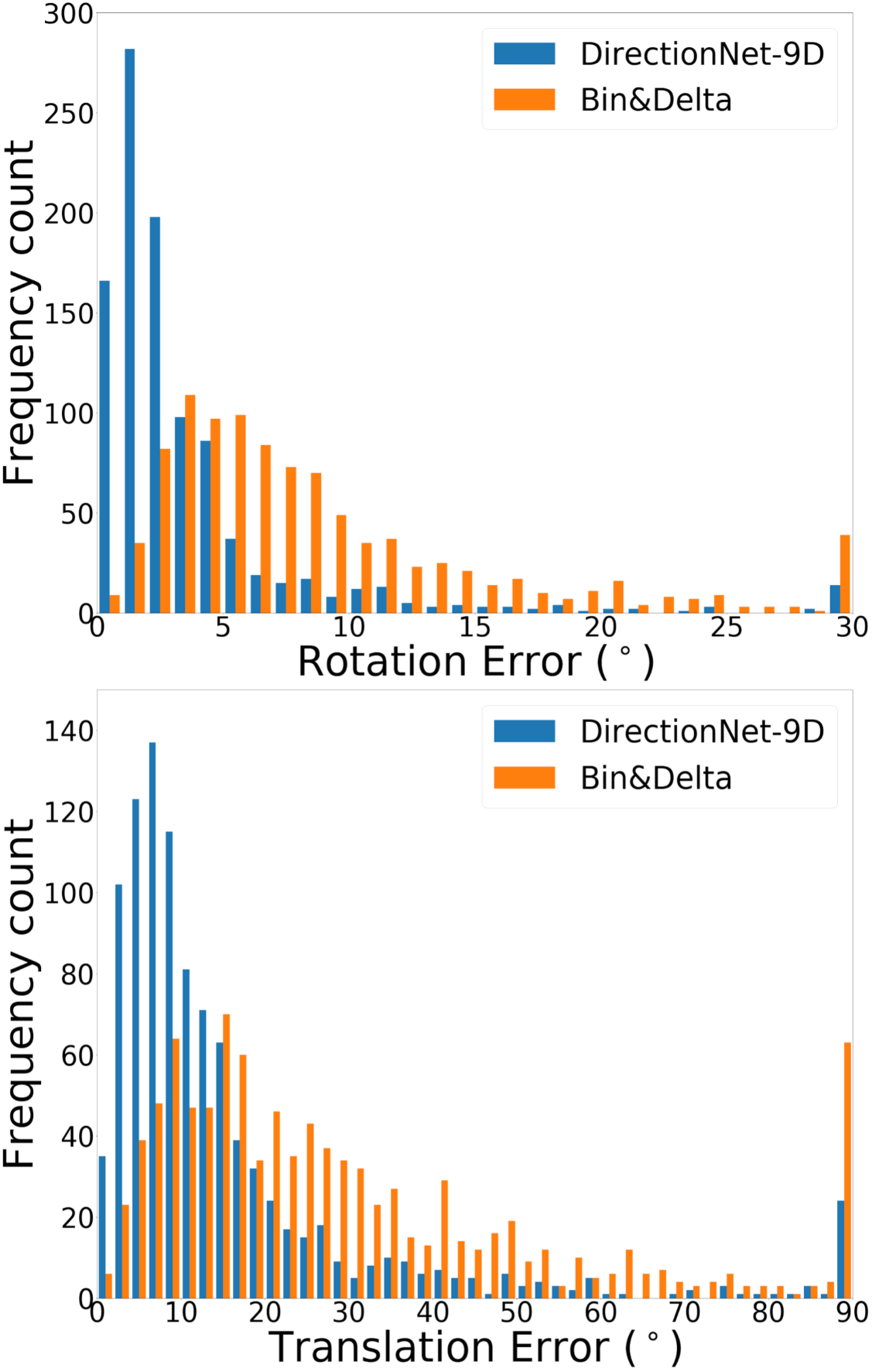}}
\hfill
\subfigure[\MatterportA]{\includegraphics[width=0.22\textwidth]{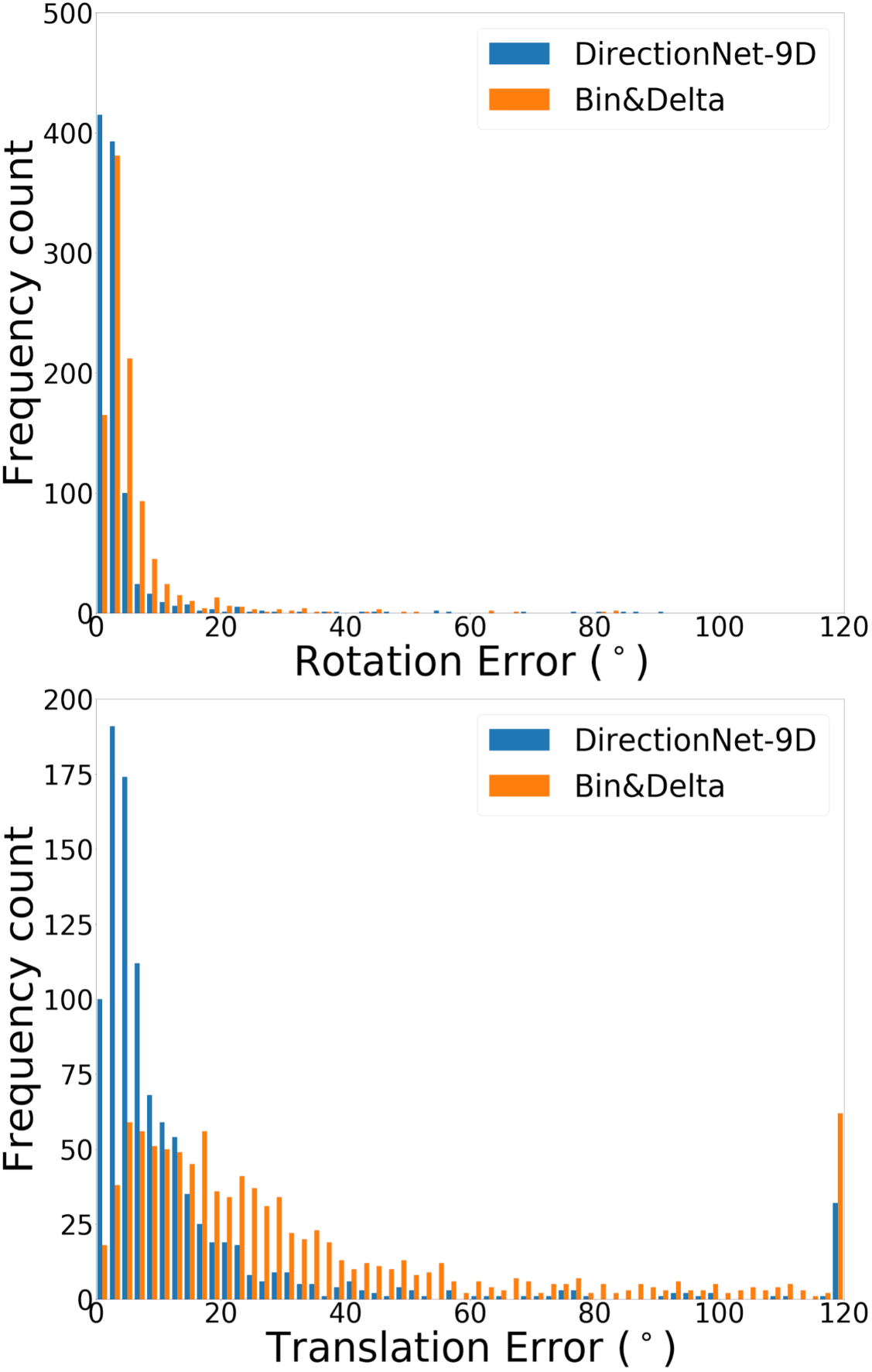}}
\hfill
\subfigure[\MatterportB]{\includegraphics[width=0.22\textwidth]{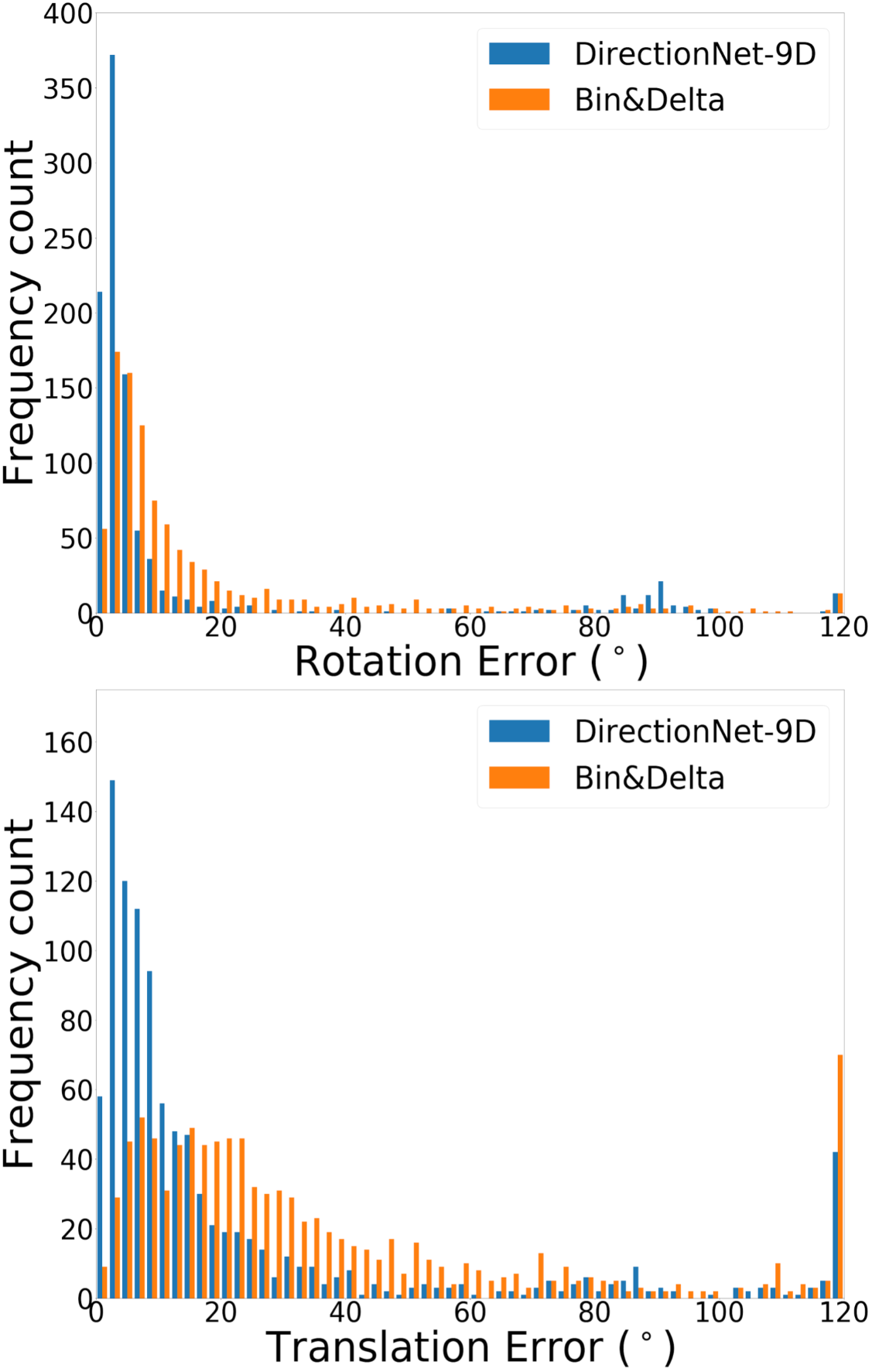}}
\caption{\textbf{Error histograms DirectionNet-9D vs. Bin\&Delta.} Top: rotation, bottom: translation.}
\label{fig:OursVsBinDelta}
\end{figure*}

\begin{figure*}[ht]
\centering
\subfigure[\InteriorNetA]{\includegraphics[width=0.22\textwidth]{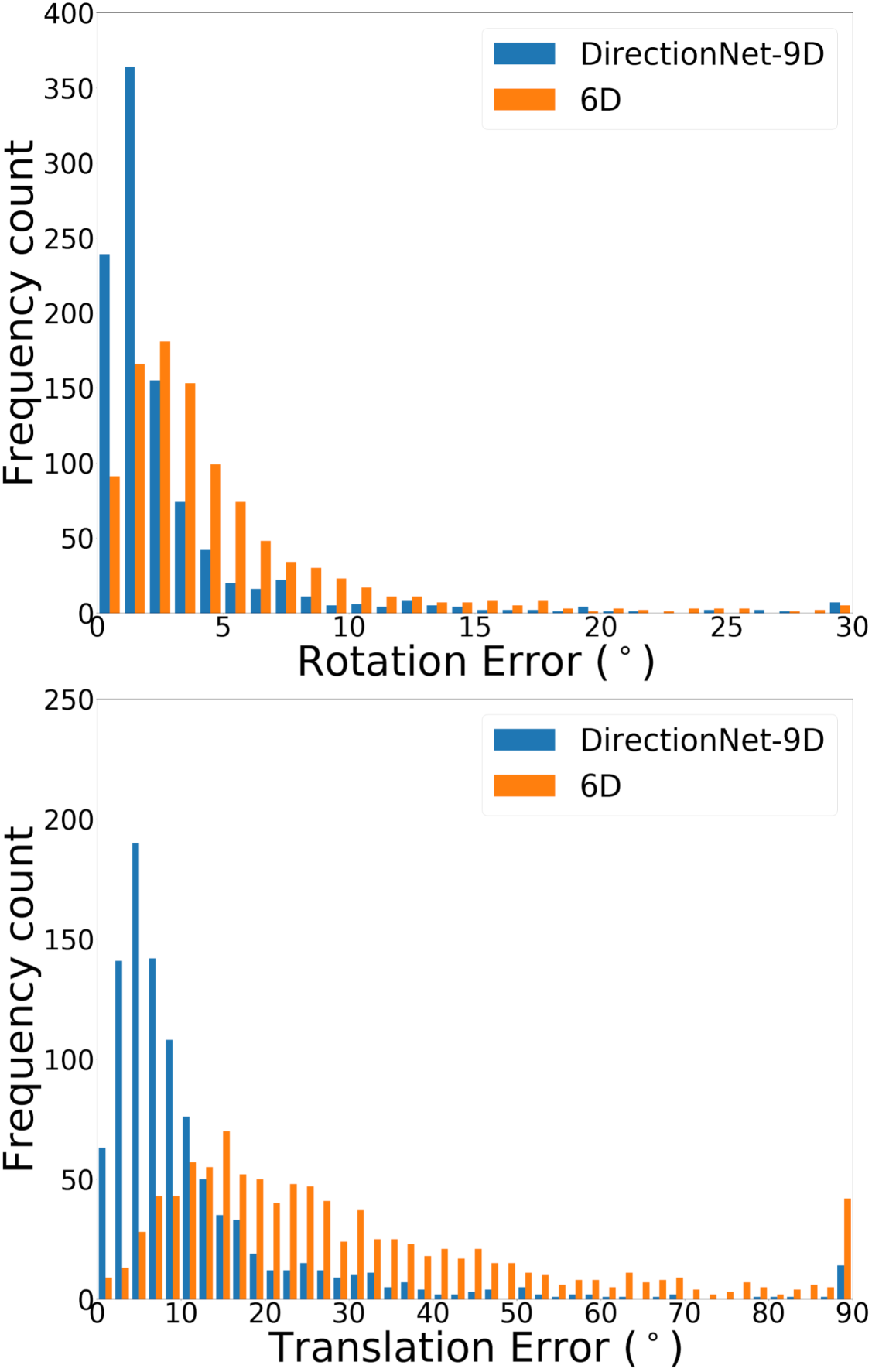}}
\hfill
\subfigure[\InteriorNetB]{\includegraphics[width=0.22\textwidth]{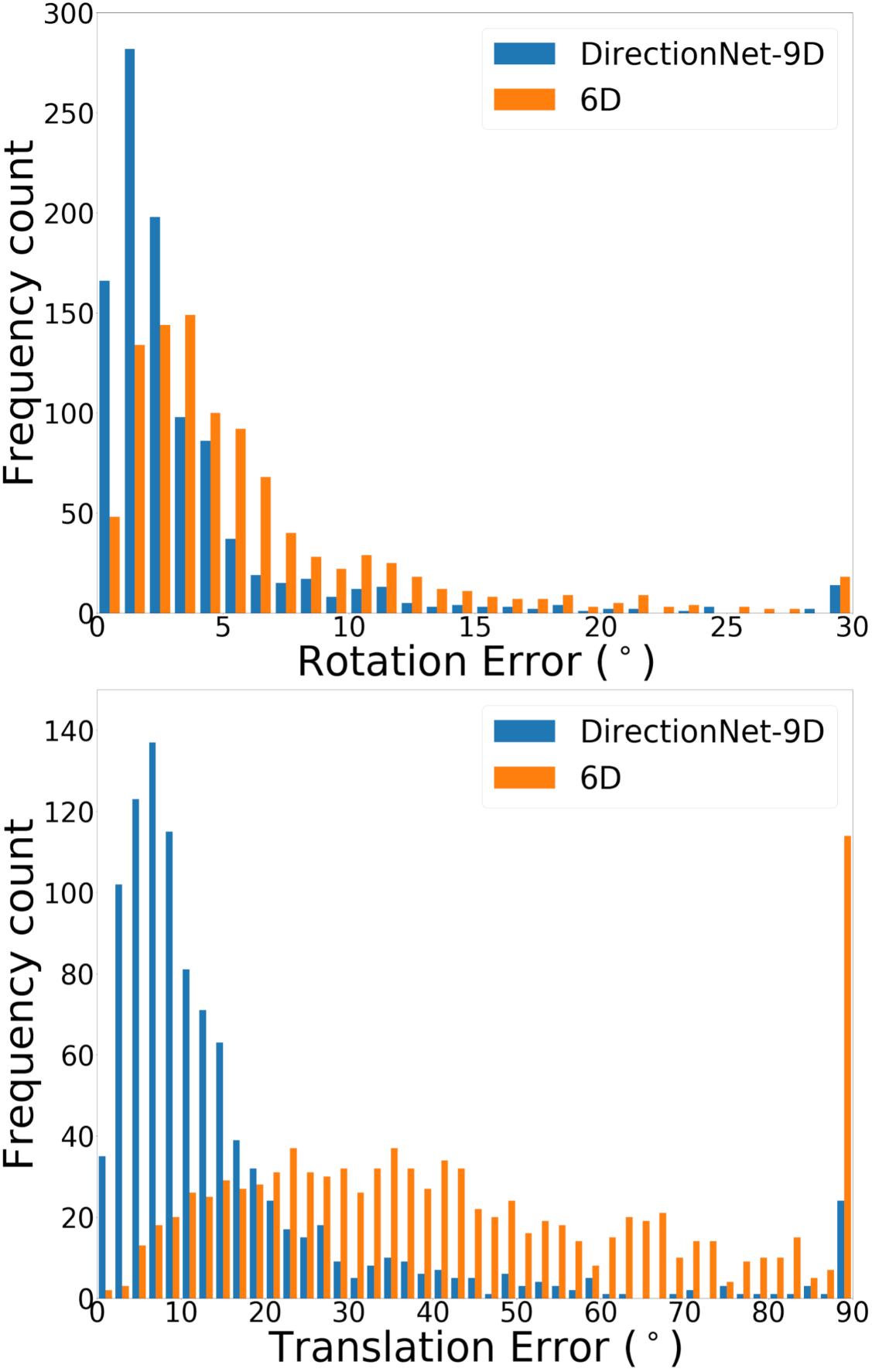}}
\hfill
\subfigure[\MatterportA]{\includegraphics[width=0.22\textwidth]{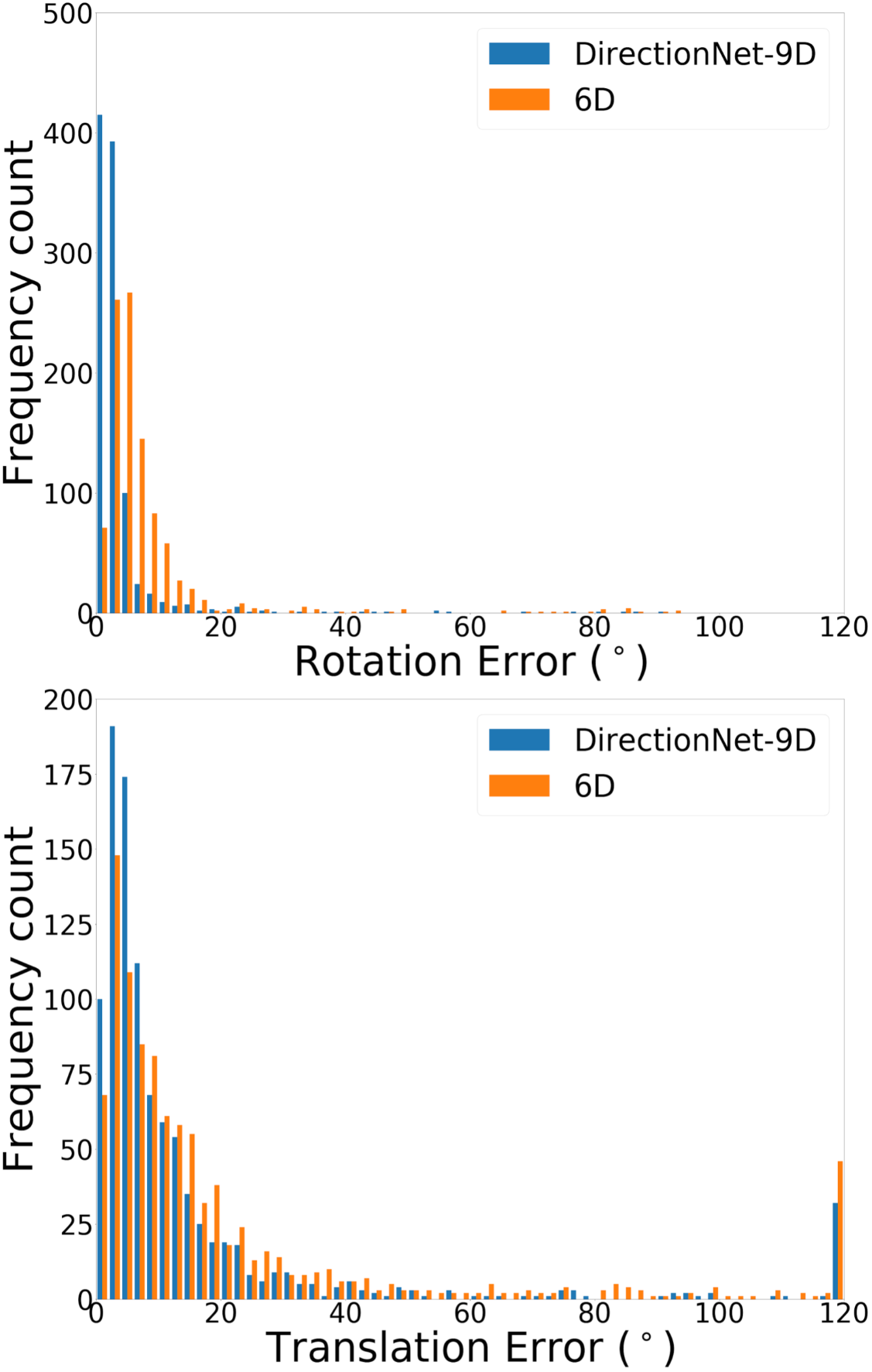}}
\hfill
\subfigure[\MatterportB]{\includegraphics[width=0.22\textwidth]{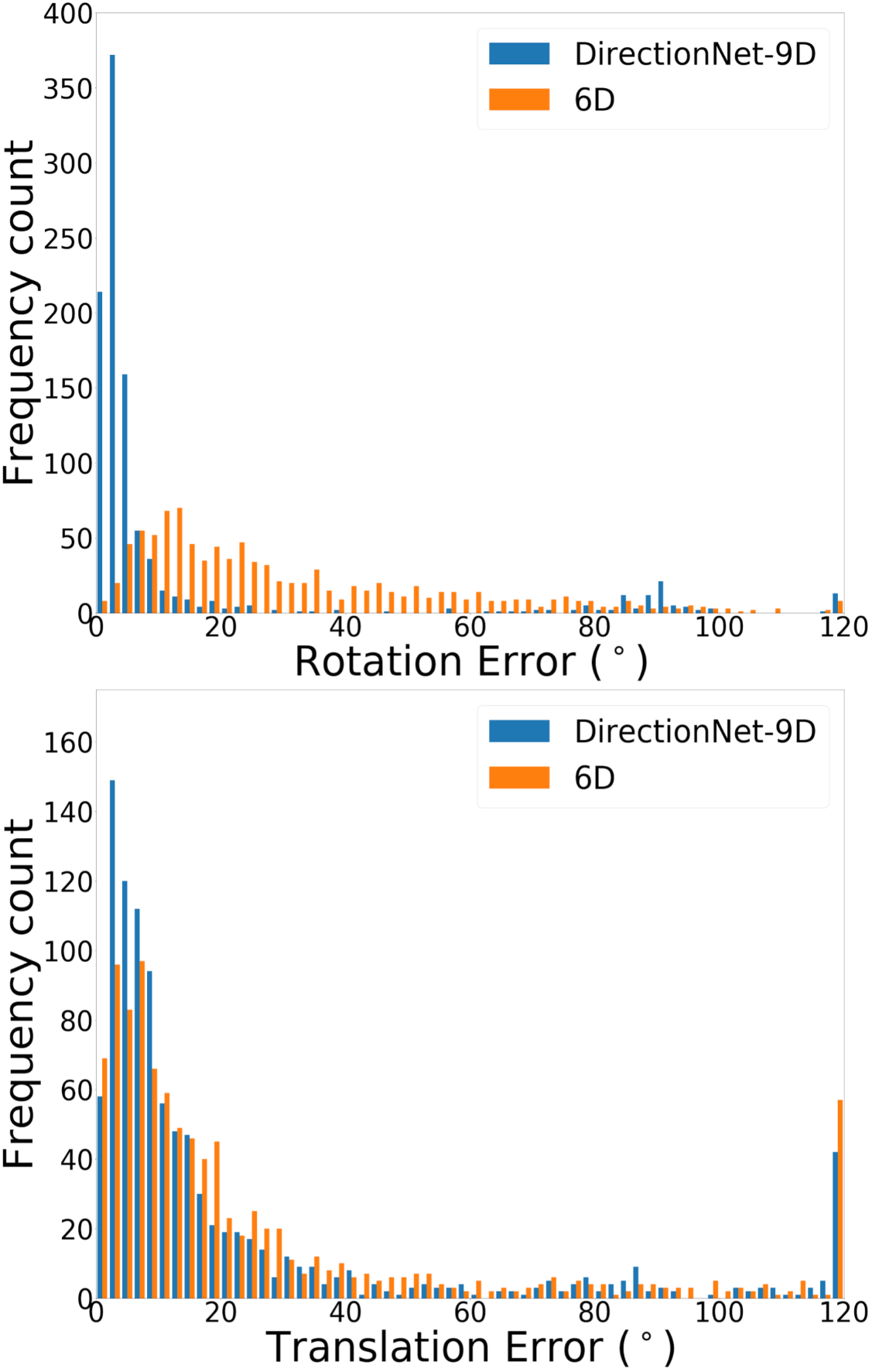}}
\caption{\textbf{Error histograms DirectionNet-9D vs. 6D Regression.} Top: rotation, bottom: translation.}
\label{fig:OursVs6DoFErrorHistogram}
\end{figure*}

\begin{figure*}[ht]
\centering
\subfigure[\InteriorNetA]{\includegraphics[width=0.22\textwidth]{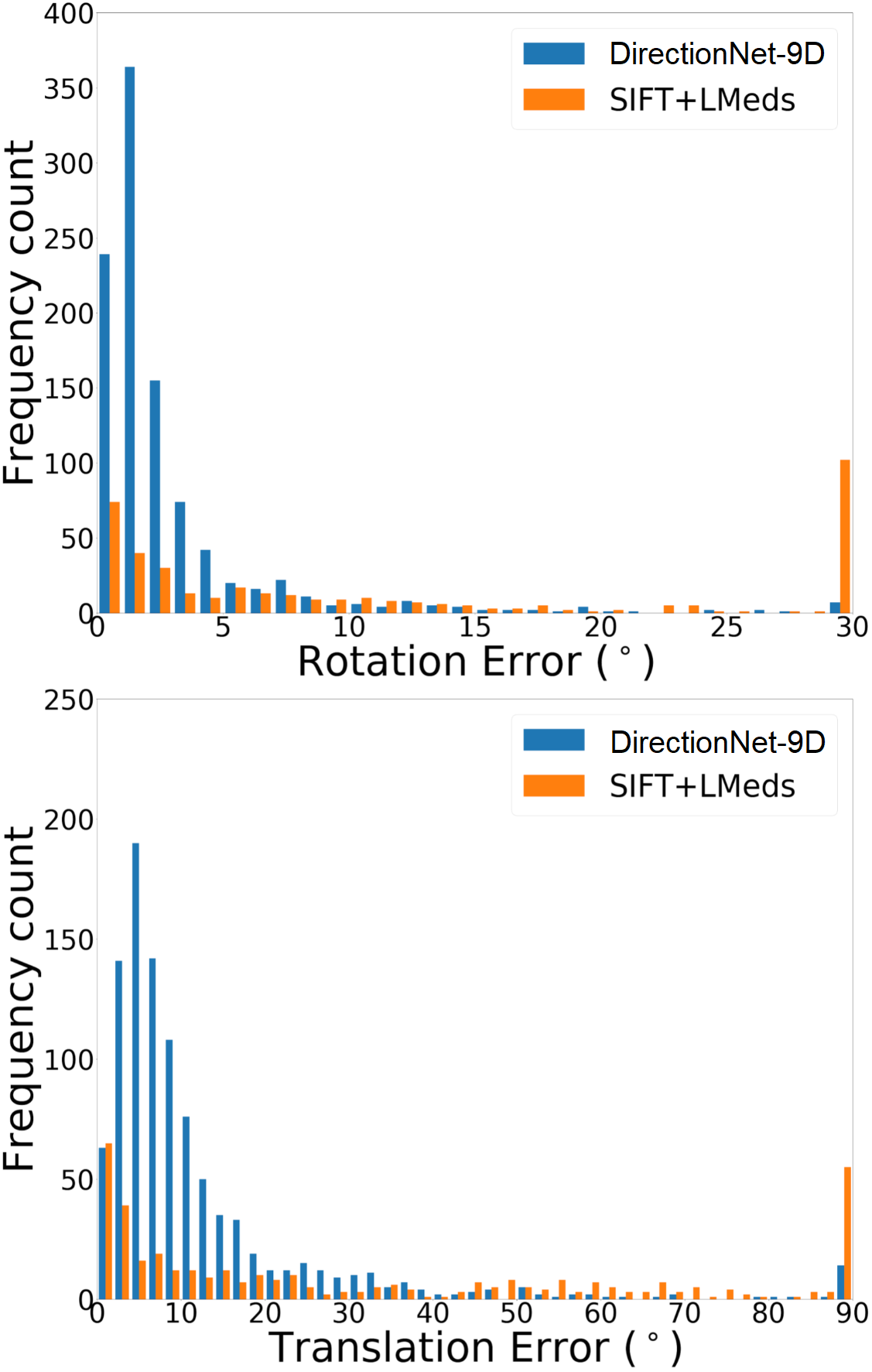}}
\hfill
\subfigure[\InteriorNetB]{\includegraphics[width=0.22\textwidth]{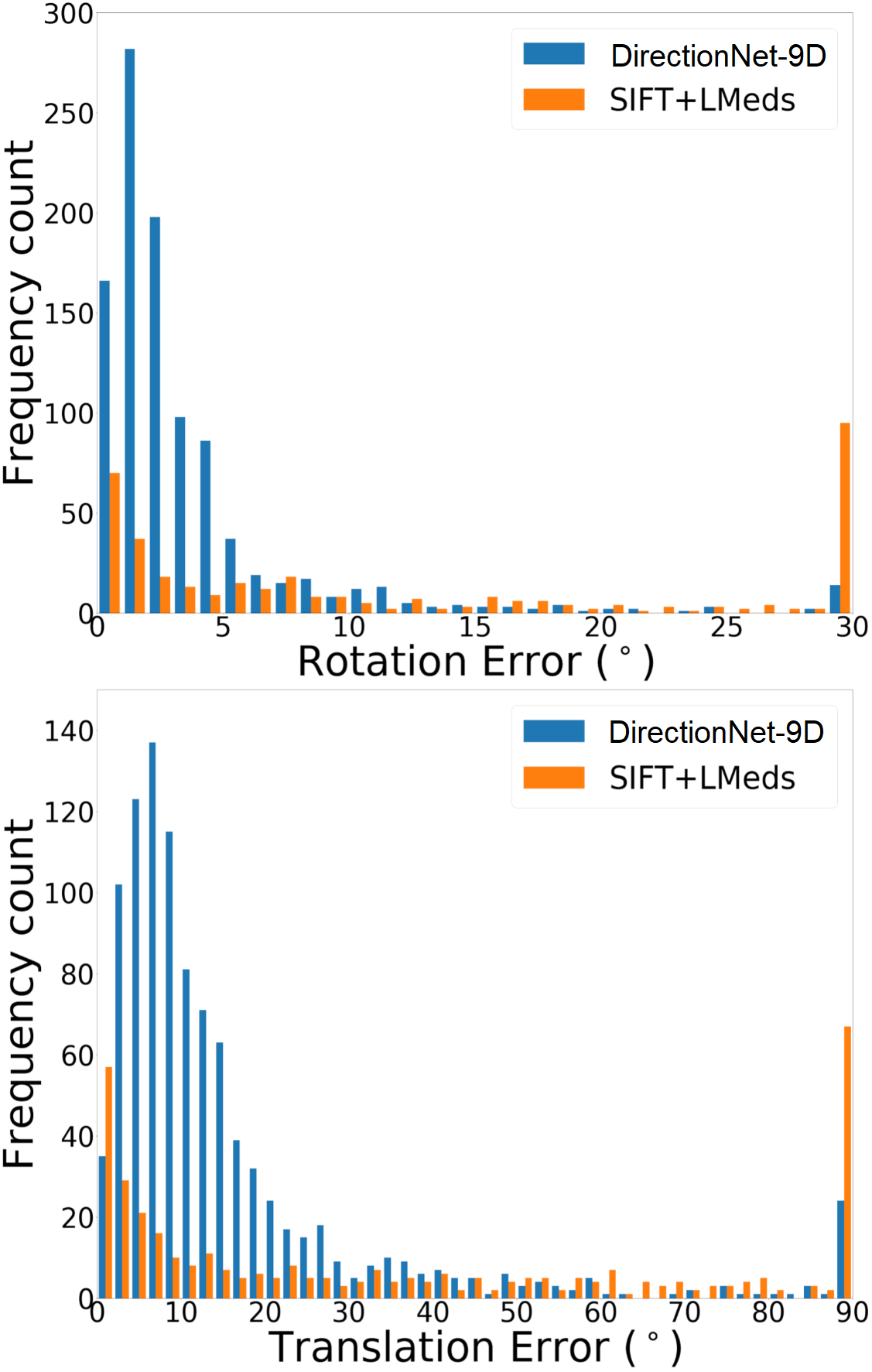}}
\hfill
\subfigure[\MatterportA]{\includegraphics[width=0.22\textwidth]{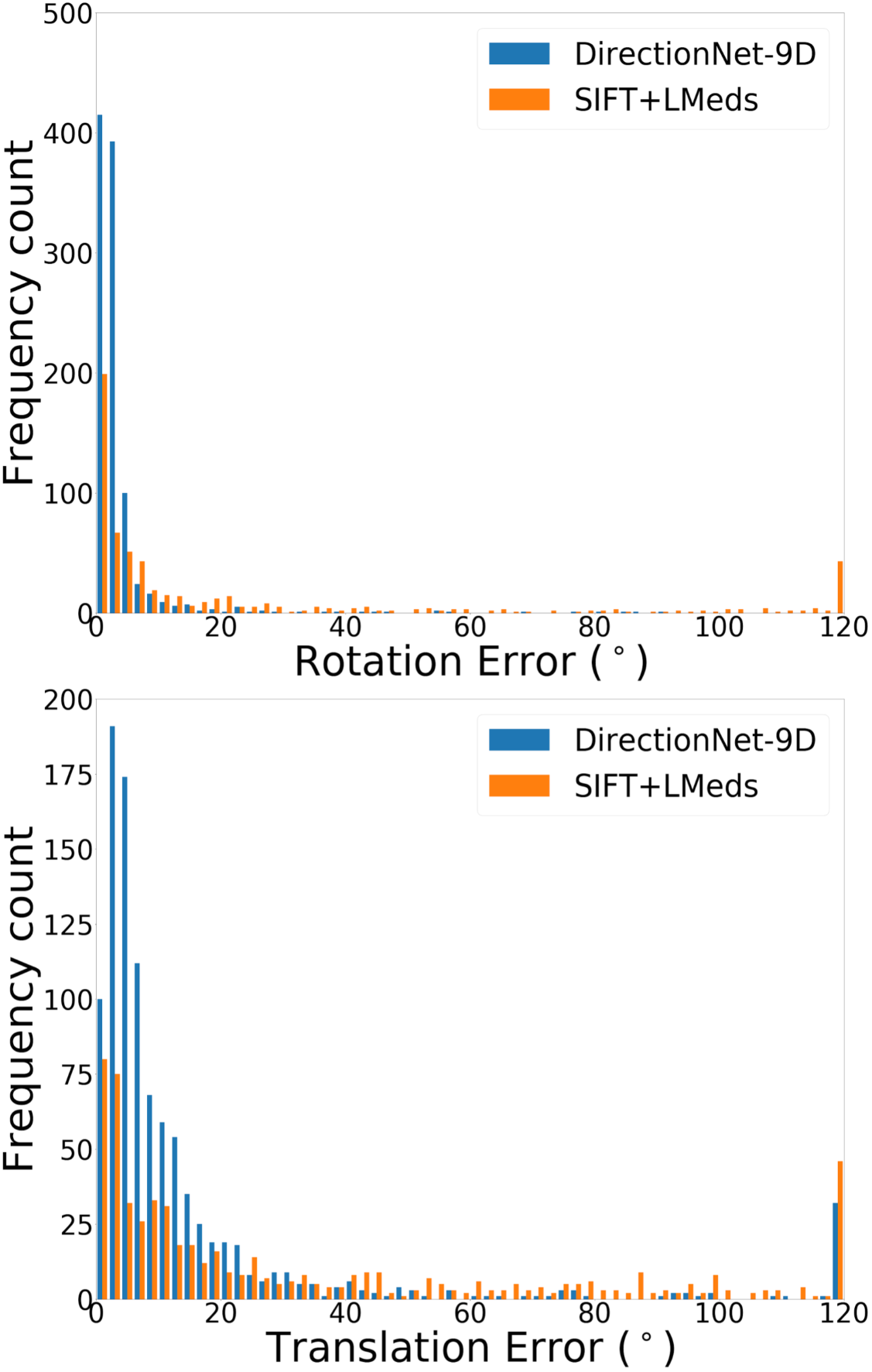}}
\hfill
\subfigure[\MatterportB]{\includegraphics[width=0.22\textwidth]{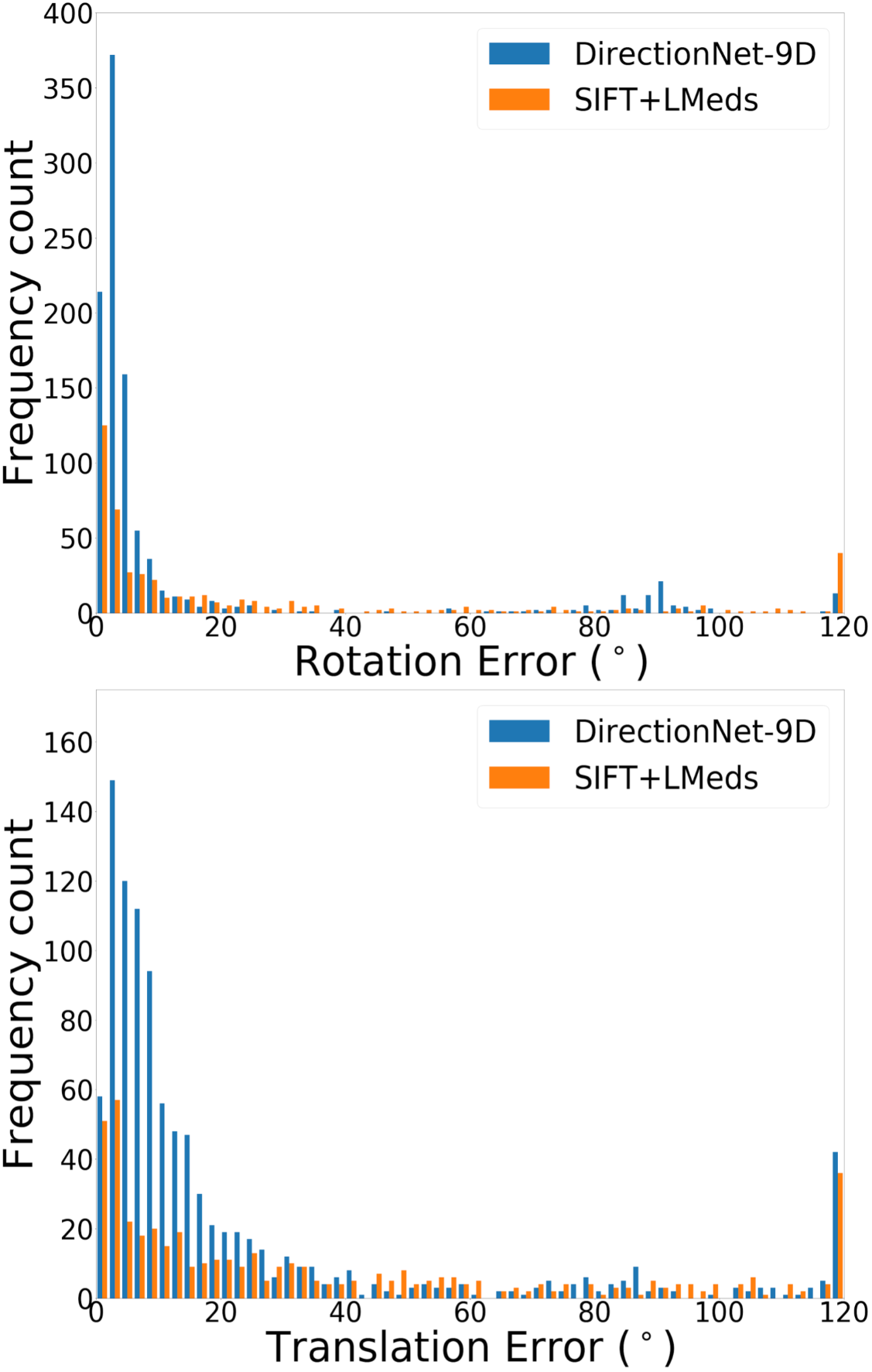}}
\caption{\textbf{Error histograms DirectionNet-9D vs. SIFT+LMedS.} Top: rotation, bottom: translation.}
\label{fig:OursVsSIFT}
\end{figure*}

DirectionNet consistently outperforms regression methods, showing the potential value in a fully convolutional model that avoids fully-connected regression layers and discontinuous parameterizations of pose. We show more comprehensive results to compare our model DirectionNet-9D with the baselines. 
Note that the spherical regression baseline generally has a higher error in rotation compared with the 6D regression method. Even though the spherical exponential activation does improve training the regression model, the 6D continuous rotation representation is still preferable to quaternions. Figure~\ref{fig:OursVsBinDelta} and Figure~\ref{fig:OursVs6DoFErrorHistogram} compare the error histogram distribution of our model with the best two regression baselines, the Bin\&Delta and the 6D Regression. Figure~\ref{fig:OursVsSIFT} compare the error histogram distribution of ours with SIFT+LMedS. Note that SIFT+LMeds has a higher mode close to 0 degree error compared with other baselines. With accurate correspondences, feature-based methods will usually outperform deep learning techniques. 

\begin{figure*}[p]
\begin{center}
\includegraphics[width=0.85\textwidth]{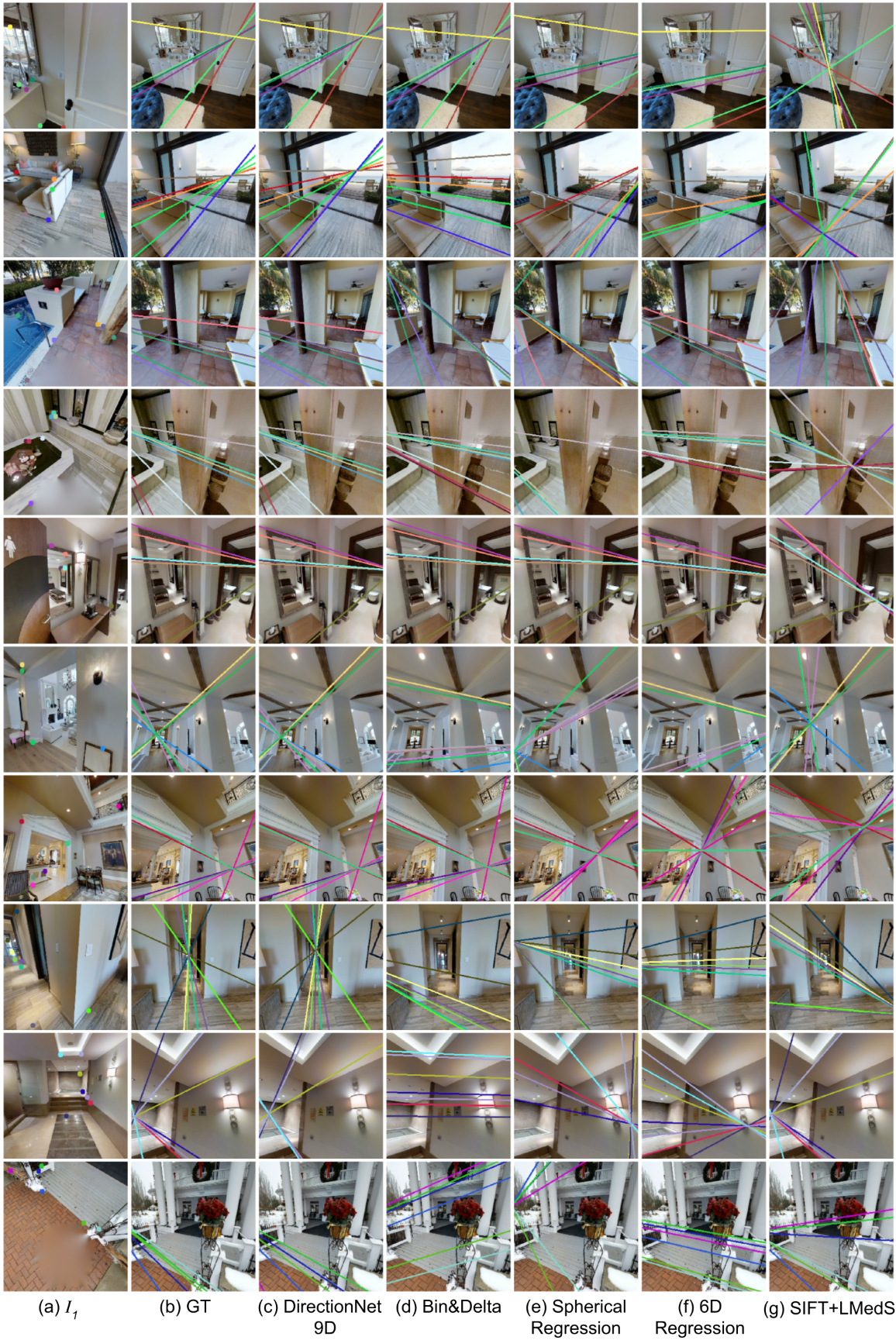}
\end{center}
   \caption{\textbf{Additional qualitative results on \MatterportB.}} 
\label{fig:MoreMatterportResults}
\end{figure*}

\begin{figure*}[p]
\begin{center}
\includegraphics[width=0.85\textwidth]{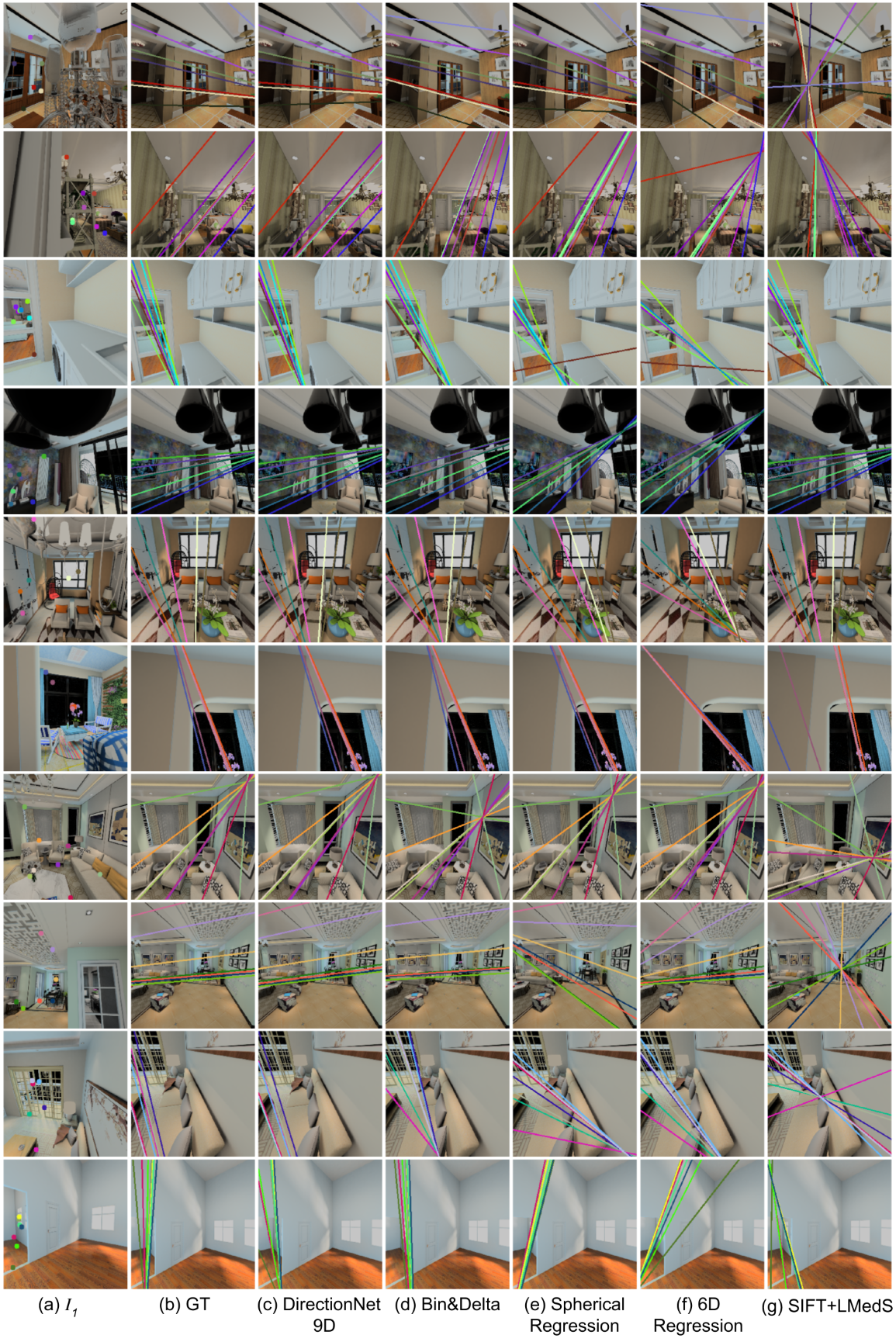}
\end{center}
   \caption{\textbf{Additional qualitative results on InteriorNet-A.}} 
\label{fig:MoreInteriorNetResults}
\end{figure*}

\begin{figure*}[p]
\begin{center}
\includegraphics[width=0.87\textwidth]{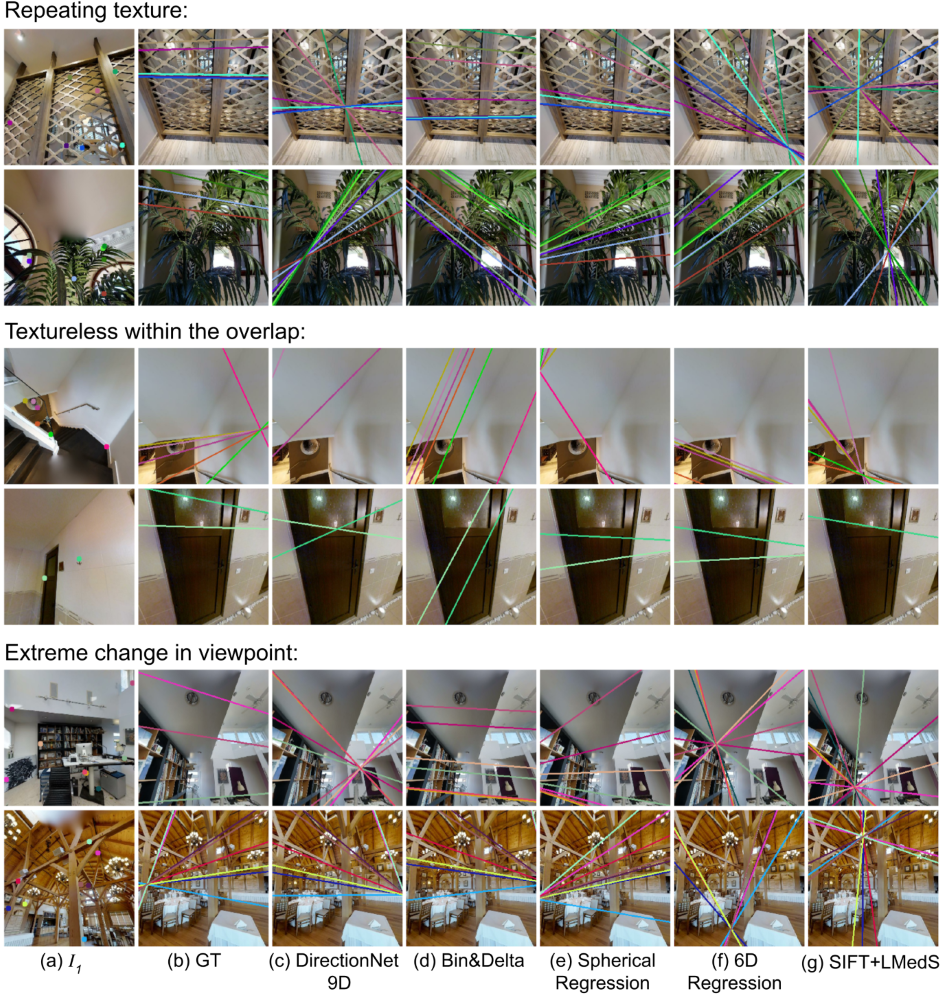}
\end{center}
   \caption{\textbf{Failure cases.} Our method could fail in cases such as repeating or complex textures, large textureless area or space with few objects, and extremely large motion.}
\label{fig:FailCases}
\end{figure*}

To visualize results of the different methods, we select a few points detected by SIFT in image $I_1$ and draw their corresponding \textit{epipolar lines} in image $I_0$ as determined by the estimated relative pose.\footnote{
Note, since we do not have ground truth point correspondences between images in our datasets, we cannot draw matching points on the two images for visualization.}  Figure~\ref{fig:EpipolarLine} illustrates the epipolar geometry. We show additional qualitative results to compare our primary model DirectionNet-9D with baselines representative of regression models and the classic method, see Figure~\ref{fig:MoreMatterportResults} and \ref{fig:MoreInteriorNetResults})  In Figure~\ref{fig:FailCases}, we highlight scenarios where our method struggles, such as repeating or complex texture, scenes with few objects and minimal texture, or extreme motion between images.

\noindent\textbf{Ablation study on loss terms.} We study the effects of the loss terms by training the DirectionNet-9D on \MatterportA. The mean rotation error is $4.68^\circ$ without the spread loss, $4.85^\circ$ without direction loss, and $14.66^\circ$ without distribution loss, compared with $3.96^\circ$ with all losses. The distribution loss which provides the direct supervision on the output distribution plays the key role in the training, because we provide the prior knowledge on the distribution by generating the ground truth from von Mises-Fisher distribution on 2-sphere which resembles the spherical normal distribution. This shows evidence that distributional learning with dense supervision is advantageous to direct regression~\cite{sun2017integral,luvizon19humanpose}. Alternatively, the distribution loss could use the KL divergence but we found MSE performs better in our experiments.

\begin{figure*}[p]
\begin{center}
\includegraphics[width=0.87\textwidth]{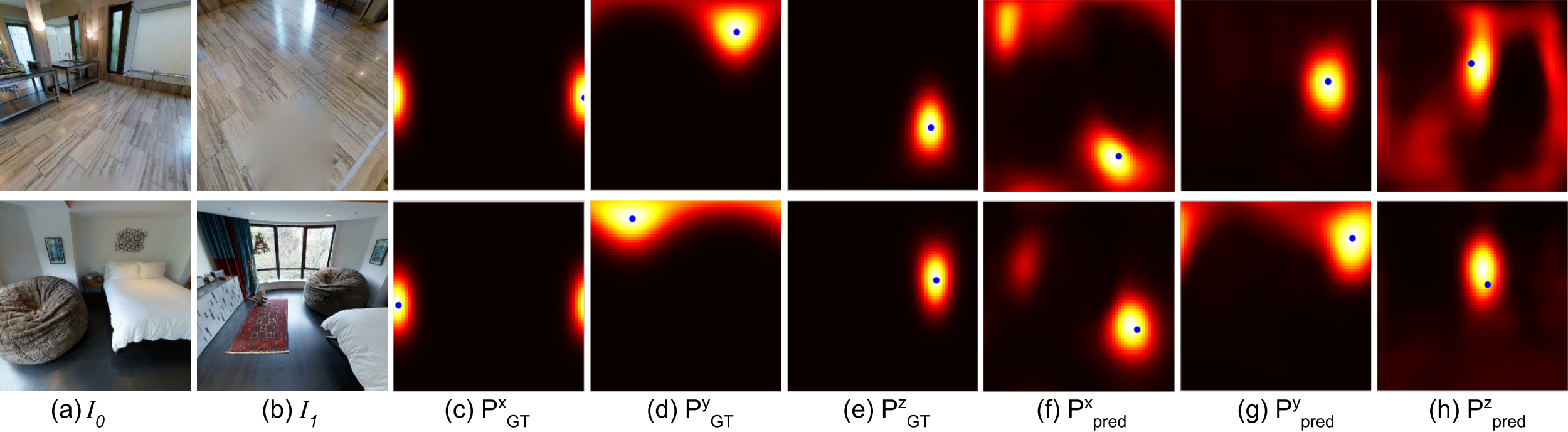}
\end{center}
   \caption{\textbf{Multimodal prediction on the rotation.} This figure shows two cases in \MatterportB when the DirectionNet-R fails and gives very high uncertainty in the presence of repeating texture or extreme motion. $I_0$ and $I_1$ are input images. ${P^x}_{GT}$, ${P^y}_{GT}$, and ${P^z}_{GT}$ are the ground truth distributions corresponds to $v_x$, $v_y$, and $v_z$ respectively. ${P^x}_{pred}$, ${P^y}_{pred}$, and ${P^z}_{pred}$ are the predictions corresponds to the three directions. The spherical distributions are illustrated as equirectangular heatmaps and the blue dot shows where the spherical expectation locates.
   } 
\label{fig:Multimodal_R}
\end{figure*}

\begin{figure*}[p]
\begin{center}
\includegraphics[width=0.75\textwidth]{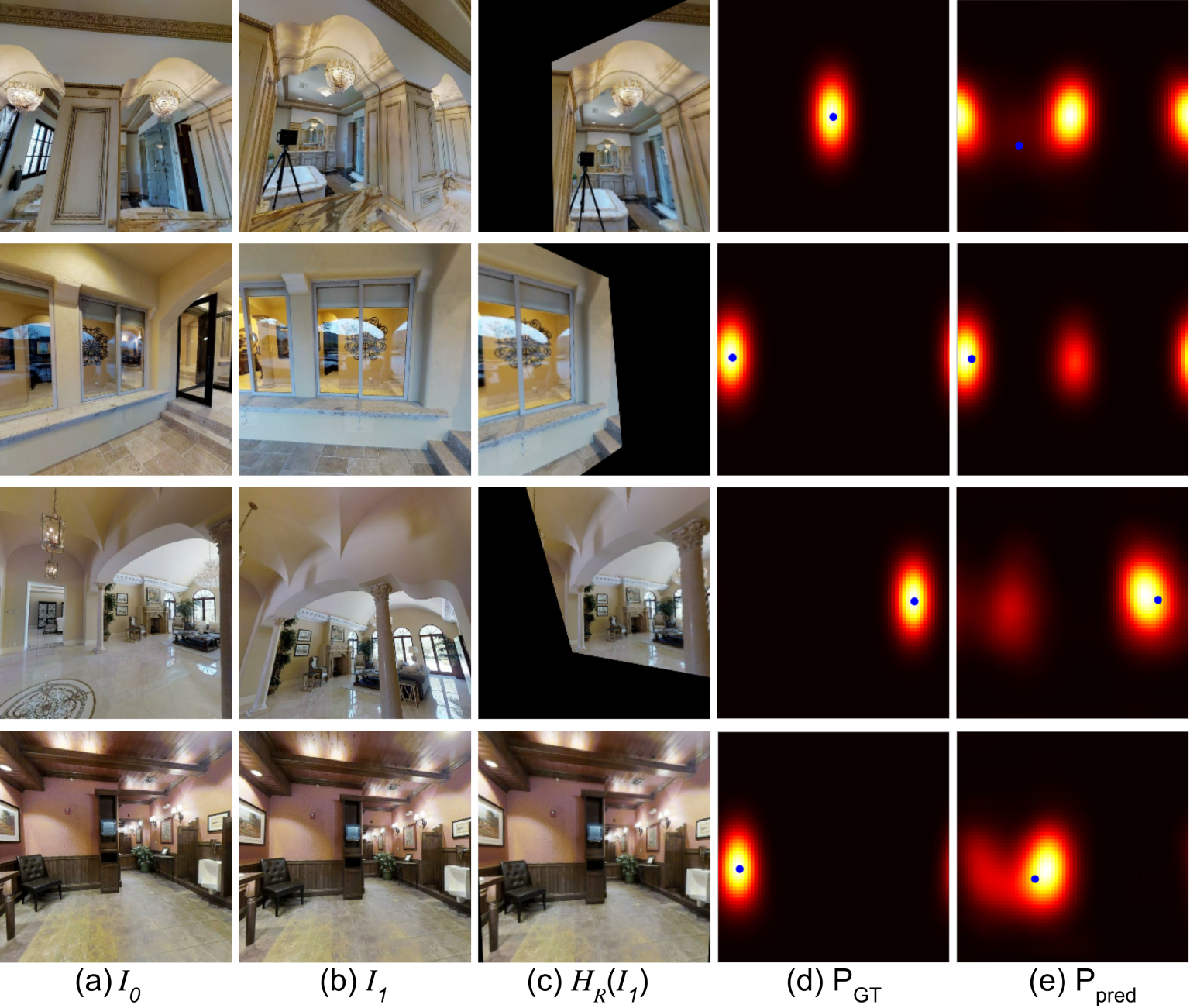}
\end{center}
   \caption{\textbf{Multimodal prediction on the translation.} This figure shows examples in \MatterportB when the DirectionNet-T produces multimodal distributions. $I_0$ and $I_1$ are original images and $H_{R}(I_1)$ is the input derotated image. ${P}_{GT}$ is the ground truth distribution of the translation and ${P}_{pred}$ is the prediction. The arcade scene in the first example has a symmetry, so it is ambiguous whether the camera is moving left or right. Thus, our model produces an antipodal distribution with almost equal uncertainty. The second example has similar symmetry because of the two identical glass windows, but we can figure out the motion from some inconspicuous clues such as the objects through the window. Thus, the model produces two modes but it is much more certain toward the correct one and the predicted direction is very close to the truth. The last two examples demonstrate another difficult scenario for the model to figure out the translation direction when the translation amount is tiny (3rd: large R and small T, 4th: tiny R and small T). The 3rd example is less ambiguous than the 4th, so the network gives high certainty at the correct mode. Note that even though the model is highly uncertain in the 4th case, the network still manages to produce a secondary mode at the correct location.} 
\label{fig:Multimodal_T}
\end{figure*}

\noindent\textbf{Multimodal distribution on high uncertainty scenarios.} In rare scenarios, our model gives higher uncertainty and produces multimodal or even antipodal distributions. Based on our observations, this usually happens in certain scenes, for example, the scene structure exhibits some symmetry or repetitive textures and causes ambiguity in the direction of the motion from two images. (See Figure~\ref{fig:Multimodal_R} and Figure~\ref{fig:Multimodal_T} for more examples.)

\noindent\textbf{Outdoor scenes.} We used KITTI odometry~\cite{Geiger2012CVPR} dataset (sequence 0-8 for train, 9-10 for eval) and sampled image pairs with a min rotation of 15$^\circ$ and translation of 10m (36K train pairs, 1K test pairs, mean translation ${\sim}18$m). Table~\ref{tab:kitti} shows generalization from MatterportA to KITTI (we cropped Matterport images to approximate the KITTI FoV). This is a hard generalization task as the distribution of relative poses in KITTI is extremely different from Matterport, yet fine-tuning with just 20\% of data is on par with the local feature baselines, and strong results after retraining with 100\% of the data indicates DirectionNet is also effective outdoors.
\begin{table}[t!]
  \centering

  \begin{adjustbox}{width=1.0\linewidth}
  \begin{tabular}{lccccc}
    \toprule

    & R mean ($^\circ$) & R med ($^\circ$) & & T mean ($^\circ$) & T med ($^\circ$) \\
    \midrule
    DirectionNet-9D (20\%) & 10.50 & 9.21 & & 26.74 & 15.67 \\
    DirectionNet-9D (100\%) & \textbf{9.19} & 6.31 & & \textbf{19.36} & \textbf{11.71} \\
    \midrule
    Regression 6D (100\%) & 13.44 & 12.74 & & 22.53 & 16.68 \\
    \midrule
    SuperGlue (outdoor) & 16.35 & 11.53 & & 24.24 & 17.15\\
    D2-Net & 24.07 & \textbf{5.18} & & 34.36 & 14.05\\
    \bottomrule\\
  \end{tabular}
  \end{adjustbox}
  \caption{\label{tab:kitti}Generalization to KITTI [15].
  }
\end{table}

\begin{figure*}[p]
\centering
\subfigure[Visualizations of RANSAC vs. LMedS]{\includegraphics[width=0.55\textwidth]{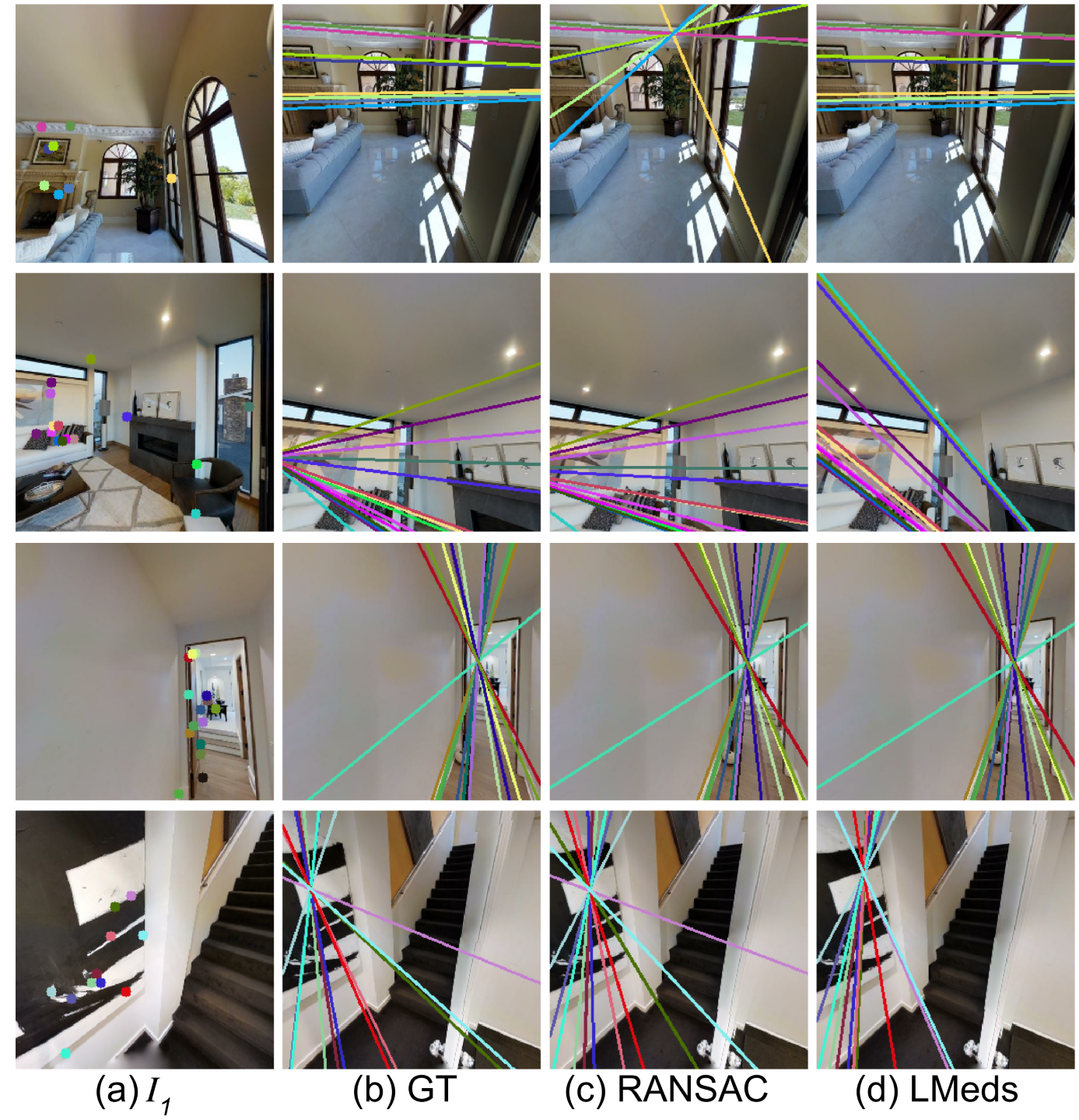}}
\hfill
\subfigure[Error histogram on \MatterportA.]{\includegraphics[width=0.35\textwidth]{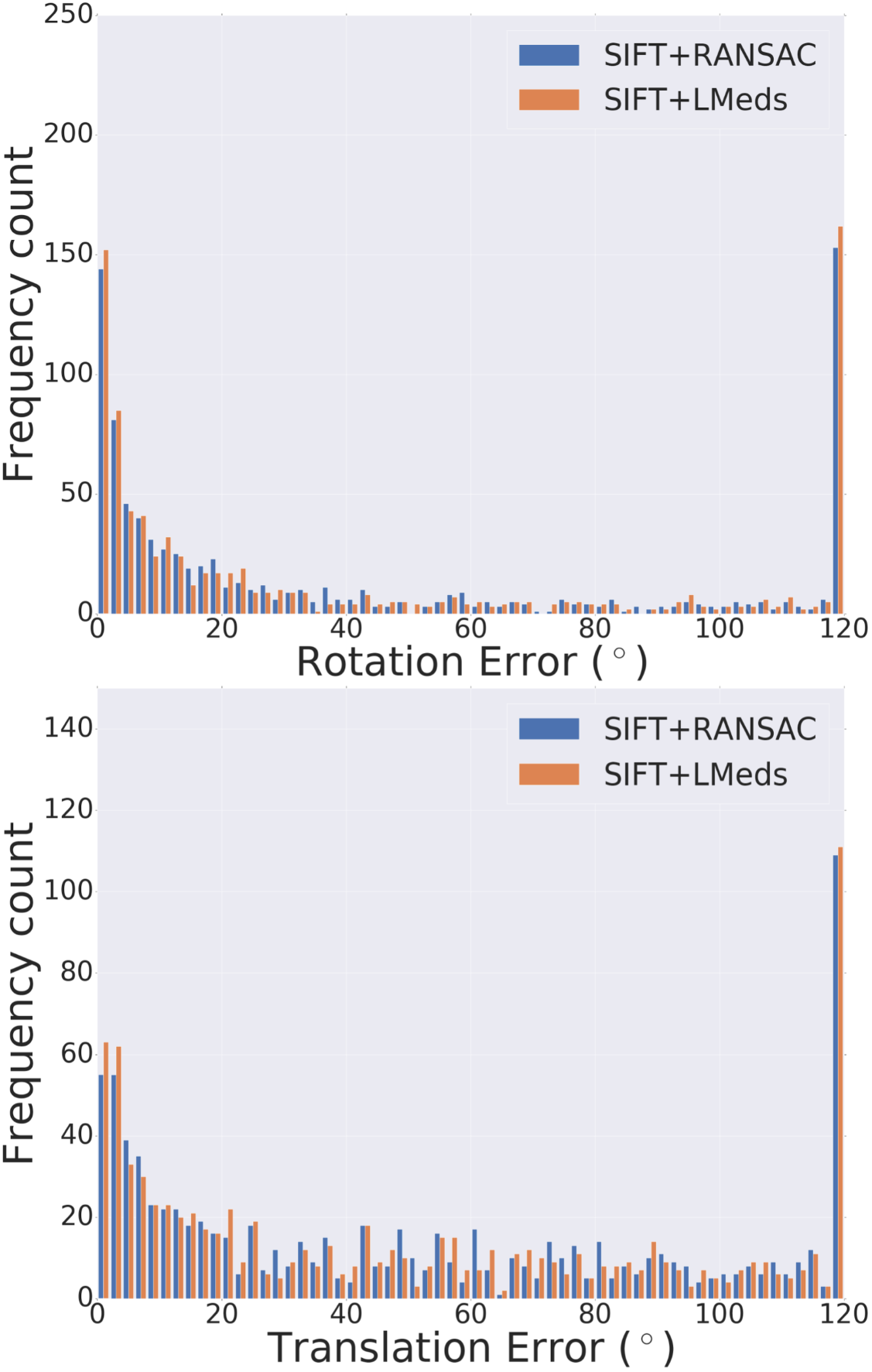}}
\caption{(a) We visualize a few examples and compare RANSAC with LMedS in different scenes. Note that the feature detected by SIFT often occurs co-planar due to the nature of the indoor scenes. (b) The error histogram shows that RANSAC and LMedS has similar performance on our dataset.}
\label{fig:RANSACvsLMeds}
\end{figure*}

\noindent\textbf{RANSAC vs. LMedS.} We use the OpenCV library (findEssentialMat() and recoverPose()) to implement both baselines by solving the essential matrix using the 5-point algorithm \cite{fivepoint} from which we recover the pose. In the main paper, we showed that LMedS performs better than RANSAC in terms of errors in translation and median errors in rotation, but RANSAC has much lower mean errors in rotation on all datasets. Note that due to the nature of indoor images, a large portion of the feature correspondences may be co-planar (e.g. features on a wall or a floor).  For RANSAC, we use the default parameters (threshold equals 1.0 and the confidence equals 0.999). Figure~\ref{fig:RANSACvsLMeds} shows that the design choice of robust fitting method doesn't make a big overall difference in our experiments.

\noindent\textbf{Runtime performance.} DirectionNet-Single inference takes under 0.02 seconds with a TESLA P100.

\end{document}